\documentclass[sigconf]{aamas} 

\usepackage{balance} 

\usepackage{algorithmic}
\usepackage{algorithm2e}
\usepackage{xcolor}
\usepackage{newfloat}
\usepackage{listings}
\usepackage{amsmath}
\usepackage{graphicx}
\usepackage{pbox}
\usepackage{subfig}
\usepackage{array, booktabs}

\setcopyright{ifaamas}
\acmConference[AAMAS '23]{Proc.\@ of the 22nd International Conference
on Autonomous Agents and Multiagent Systems (AAMAS 2023)}{May 29 -- June 2, 2023}
{London, United Kingdom}{A.~Ricci, W.~Yeoh, N.~Agmon, B.~An (eds.)}
\copyrightyear{2023}
\acmYear{2023}
\acmDOI{}
\acmPrice{}
\acmISBN{}

\acmSubmissionID{???}

\title[Follow your Nose]{Follow your Nose: Using General Value Functions for Directed Exploration in Reinforcement Learning}

\author{Durgesh Kalwar}
\affiliation{
  \institution{TCS Research}
  \city{Mumbai}
  \country{India}}
\email{durgesh.kalwar@tcs.com}

\author{Omkar Shelke}
\affiliation{
  \institution{TCS Research}
  \city{Mumbai}
  \country{India}}
\email{shelke.omkar@tcs.com}

\author{Somjit Nath}
\affiliation{
  \institution{TCS Research}
  \city{Mumbai}
  \country{India}}
\email{somjit.nath@tcs.com}

\author{Hardik Meisheri}
\affiliation{
  \institution{TCS Research}
  \city{Mumbai}
  \country{India}}
\email{hardik.meisheri@tcs.com}

\author{Harshad Khadilkar}
\affiliation{
  \institution{TCS Research}
  \city{Mumbai}
  \country{India}}
\email{harshad.khadilkar@tcs.com}

\begin{abstract}
Improving sample efficiency is a key challenge in reinforcement learning, especially in environments with large state spaces and sparse rewards. In literature, this is resolved either through the use of auxiliary tasks (subgoals) or through clever exploration strategies. Exploration methods have been used to sample better trajectories in large environments while auxiliary tasks have been incorporated where the reward is sparse. However, few studies have attempted to tackle both large scale and reward sparsity at the same time.
This paper explores the idea of combining exploration with auxiliary task learning using General Value Functions (GVFs) and a directed exploration strategy. We present a way to learn value functions which can be used to sample actions and provide directed exploration. Experiments on navigation tasks with varying grid sizes demonstrate the performance advantages over several competitive baselines. 
\end{abstract}


\newcommand{\BibTeX}{\rm B\kern-.05em{\sc i\kern-.025em b}\kern-.08em\TeX}
\DeclareMathOperator{\E}{\mathbb{E}}

\begin{document}
\RestyleAlgo{boxruled}

\pagestyle{fancy}
\fancyhead{}


\maketitle 


\section{Introduction}

Reinforcement Learning (RL) has shown great promise in recent years for solving sequential decision making tasks. Particularly in the field of games, Reinforcement Learning algorithms have managed to achieve superhuman levels of performance\cite{DBLP:journals/corr/MnihKSGAWR13} \cite{SilverHuangEtAl16nature}. A recurring theme in the design of RL algorithms is the exploration vs exploitation tradeoff, particularly in hard exploration environments. It is a common practice to use $\epsilon$-greedy strategies for exploration. However, for hard exploration environments this strategy is unable to learn efficiently due to the low probability of generating sensible trajectories with purely random exploration. This property has been observed in environments such as the game Montezuma's Revenge in Atari \cite{DBLP:journals/corr/MnihKSGAWR13}.

The main goal in RL tasks is to obtain an optimal policy that maximises the expected discounted return (with discount factor, $\gamma \in [0,1]$). The optimal policy, $\pi^{*}$ can be defined as:
$$
    \pi^{*} = \underset{\pi}{\mathrm{argmax}}\, \E_{\pi} [\sum_{t=0}^{\infty} \gamma^{t}r_t(s_t,a_t)]
$$
from interaction with the environment in the form of states $s_t$, actions $a_t$, and rewards $r_t$. At the beginning, the agent will not have much idea about the environment it is in and thus needs to gather information before it can take useful actions. If the state space is very complicated and/or the reward is very sparse, obtaining useful trajectories (collection of states, actions and rewards) can be difficult. This is the problem of exploration in RL which is still quite an open area for research. 

There have been many recent papers that address the problem of exploration like Noisy Networks \cite{DBLP:conf/iclr/FortunatoAPMHOG18}, Count-based Exploration \cite{DBLP:conf/nips/TangHFSCDSTA17}, Curiosity Driven Exploration \cite{DBLP:conf/icml/PathakAED17}, unsupervised learning of goal space for intrinsically motivated exploration \cite{DBLP:conf/iclr/PereFSO18}, Go-explore \cite{DBLP:journals/corr/abs-1901-10995} which goes to a previously explored state and restarts exploration from that state. 
One interesting breakthrough for solving hard exploration problems came in the Never Give Up \cite{DBLP:conf/iclr/BadiaSVGPKTAPBB20} paper, which was included in the Agent 57 \cite{DBLP:conf/icml/BadiaPKSVGB20} paper which is able to successfully solve all the 57 Atari games. This paper introduces two additional rewards for defining directed exploration, the episodic and the intrinsic reward. Agent 57 uses a family of policies with varying degree of exploration trained by Recurrent Experience Replay in Distributed Reinforcement Learning (R2D2) \cite{DBLP:conf/nips/RevaudSHW19} with Retrace ($\lambda$) \cite{DBLP:conf/nips/MunosSHB16} to account for off-policy learning. Most of these methods require learning some form of novelty or generating a reward function based on curiosity. None of them are as simple as the traditional $\epsilon$-greedy approach. Recently, \cite{DBLP:conf/iclr/DabneyOB21} came up with a temporally extended version of the $\epsilon$-greedy exploration strategy that can handle complicated environments as well.

As explained in \cite{DBLP:conf/iclr/DabneyOB21}, the various other exploration strategies developed keeping difficult environments in mind tend to over-fit to those challenging environments and do not work as well in environments requiring lesser exploration. The authors further explain how temporally extended epsilon greedy is a sound strategy that can work well in a variety of environments. In this paper, we aim to demonstrate the following contributions:
\begin{enumerate}
    \item We present an exploration strategy designed for large state spaces with sparse rewards. This is achieved by defining auxiliary tasks with the help of General Value Functions (GVFs) to perform directed exploration, thereby further improving state space coverage during exploration. Our strategy improves on temporally extended $\epsilon$-greedy exploration with sub-policies from the General Value Functions to perform better exploration.
    \item Extensive experiments over two sets of navigation environments, which provide empirical evidence about superior performance and robustness to hyperparameters such as the $\epsilon$ schedule.
\end{enumerate}


\section{Background}
\begin{figure}
    \centering
    \includegraphics[width=0.95\linewidth]{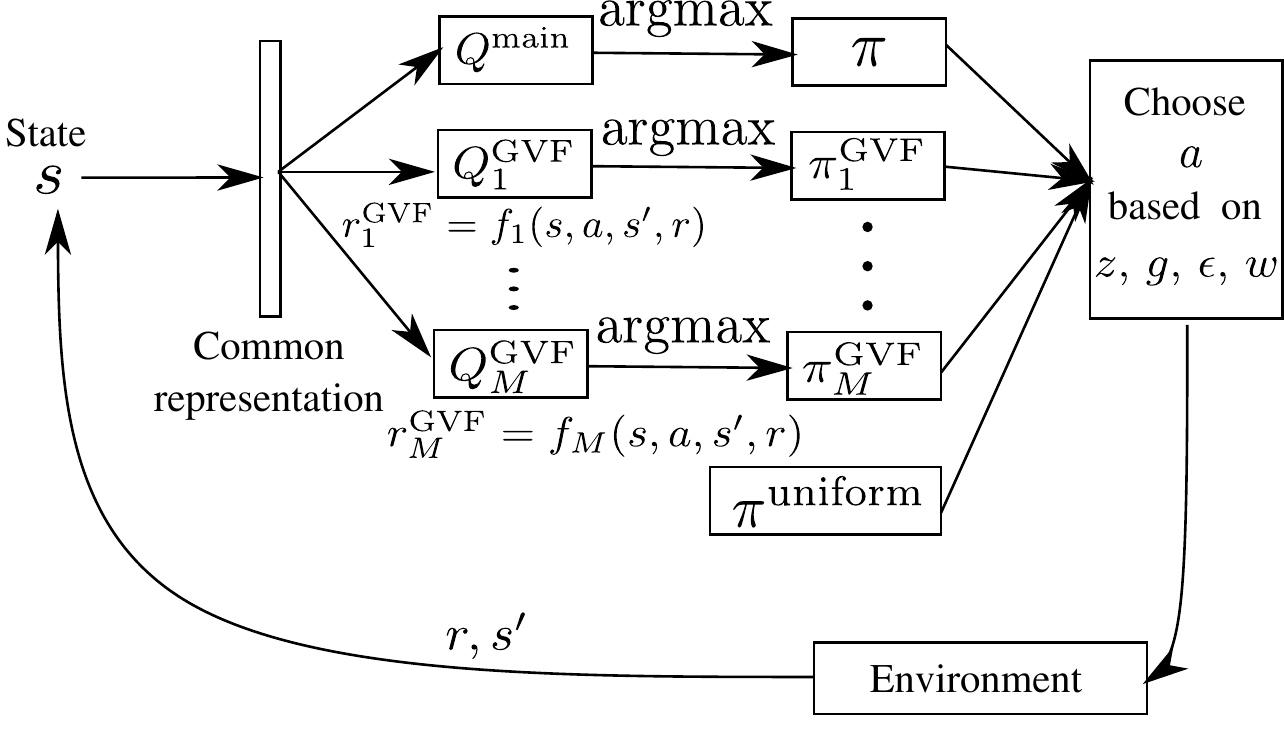}
    \caption{High-level architecture of the DEZ-exploration policy with $M$ GVFs, one main policy, and one uniform policy. The reward for training each GVF is computed using handcrafted functions $f$. The logic for choosing action $a$ is described in Algorithm \ref{alg:DEZ}.}
    \label{fig:network}
\end{figure}
\textbf{Temporally Extended EZ-Greedy Exploration}: In generic $\epsilon$-greedy exploration, the agent can either take an action that maximizes its own state-action values with probability 1-$\epsilon$ or a random action with probability $\epsilon$. Thus given infinite time, it would explore the entirety of the state space as it would be taking random actions infinite times provided we have a non-zero $\epsilon$ always. This form of exploration is generally very slow in covering different parts of the state space. Thus temporally extended $\epsilon$-greedy \cite{DBLP:conf/iclr/DabneyOB21} fixes a random action and keeps taking the action for a certain number of time steps, called persistence ($Z$) of the random action. Instead of taking one random step with probability $\epsilon$, it executes a sequence of the same random actions. The persistence $Z$ is not a constant value throughout the exploration process however it is sampled from a uniform distribution between 0 (no exploration) and $Z_{max}$ (a hyper-parameter specific to the environment). This is why the algorithm is also known as EZ-Greedy.


In the paper, the authors mostly used the same random action repeated K times, but instead, we show that sampling actions from the greedy policy of learned General Value Functions (GVFs) \cite{10.5555/2031678.2031726} K times yields better-directed exploration. This can help to reach parts of state-space that could be novel or would help in solving the main goal. In this paper, we show that having GVFs which are trained over auxiliary tasks provides an efficient way to explore as compared to count-based learning (curiosity-driven learning) or reward shaping.

\noindent \textbf{General Value Functions:} A value function provides an estimate of the discounted total expected return that can be obtained given a certain state-action pair and a policy $\pi$. Similarly, General Value Functions (GVFs) \cite{10.5555/2031678.2031726} estimate a user-specified environment property, optimizing over the user-specified reward function given a policy and a discount factor. Thus, if the discounted returns be $G_t$ and for a trajectory length T,
$$Q^{GVF}({s,a;\pi,\gamma, r) = \E[G_t | s_t,a_t,A_{t+1:T-1} \sim \pi}].$$
The input to the GVFs would thus be a policy, discount, reward function (also called cumulants). The `question' asked of a GVF is the estimate of the discounted return for the specified cumulants. For each question, GVFs can be trained with normal RL algorithms (like Q-learning) and thus can provide useful information about the environment and the transition dynamics. In the Horde architecture \cite{10.5555/2031678.2031726}, the authors show the usefulness of answering predictive questions about the environment as well as the capability of learning multiple control policies from a single behaviour policy.  

The cumulants can also be viewed as providing auxiliary tasks the agent needs to perform before it can solve the main task. Often such auxiliary tasks would have no reward signal associated with them otherwise. The tasks may be defined using handcrafted cumulants \cite{DBLP:conf/iclr/JaderbergMCSLSK17} or they may themselves be learnt on their own \cite{DBLP:conf/nips/VeeriahHXRLOHSS19}. In this paper, we define the GVF reward functions and termination functions by hand, and learn the GVF values off-policy i.e. they are learned with respect to their own greedy policies. Thus, maximising for $Q^{GVF}$ generates a different sub-task oriented policy which can be viewed as an subroutine, to be used for a short persistence period. Thus, by following the policy for a short period we can explore directed by a subpolicy.


\section{Proposed Directed EZ-Greedy Strategy}
Directed EZ-Greedy learns auxiliary tasks in the form of GVFs and uses sub-policies obtained by greedy action selection from the General Value Functions for temporally extended $\epsilon$-greedy algorithm. With probability $\epsilon$ it selects a random option which can be sub-policies from any of the randomly chosen GVFs at that particular state and keeps taking the greedy action with respect to that GVF. Thus, it explores different parts of the state space by following a specific GVF policy which makes the exploration directed, hence the name Directed EZ-Greedy. The entire algorithm is portrayed in Algorithm \ref{alg:DEZ}.

\begin{algorithm}
\caption{DEZ-Greedy exploration strategy}

\SetKwFunction{FMain}{\text{DEZGreedy}}
\SetKwProg{Fn}{\text{Function}}{:}{}
\Fn{\FMain{$\epsilon$, $Z_{max}$}}{
Countdown timer $z \leftarrow 0$ \\
Uniformly sampled random action $w \leftarrow -1$ \\ 
Selected GVF index $g \leftarrow 0$ \\
\While{\text{True}}{
    \text{Observe state}  $s$ \\
    \eIf{$z==0$}
    {
        // Timer ran out: choose controlling policy\\
        \eIf{$random() < \epsilon$}{ // Explore \\
        \text{Sample duration:} $z \sim [1, Z_{max}]$ \\
        \text{Sample GVF:} $g \sim [0,M]$ \\
        \eIf{$g==0$}
        {\text{Sample action:} $w \leftarrow U(A)$ \\ $a \leftarrow w$}
        {$a \leftarrow argmax(Q^{GVF}_g)$}
        }
        {// Exploit \\ $a \leftarrow argmax(Q^{Main})$}
    }
    {
        // Continue previous exploration policy \\
        \eIf{$g==0$}{$a \leftarrow w$}{$a \leftarrow argmax(Q^{GVF}_g)$}
        $z \leftarrow z-1$
    }
    \text{Take action} $a$
    }
}

\label{alg:DEZ}
\end{algorithm}

The algorithm learns M+1 Value Functions as shown in Figure \ref{fig:network} where M is the number of GVFs. These Value Functions are learned from a shared common representation through which the losses for both the main Q values and the GVFs propagate. In Algorithm \ref{alg:DEZ}, we sample an action from M+1 ([0, M]) possible values. In all our experiments, we include the option of repeating a random action as one additional option as well. This is added to make sure that theoretically, it is still possible to visit all states in the state space and we have asymptotic full state space coverage.

There are two main advantages of this approach:
\begin{enumerate}
    \item GVFs have been generally used to give predictive knowledge of the environment and this directly helps in learning good representations. In Figure \ref{fig:network}, we also employ a similar structure whereby the GVFs not only influence the exploration policies but also help in learning better representations.
    \item Learning GVFs while following the behavior policy leads to divergence if we do not use importance sampling ratios as the greedy policy of the GVFs will be off-policy. However following the GVF policy initially helps to keep the behavior policy closer to the greedy GVF policy (at least during the exploration phase). 
\end{enumerate}

\begin{figure}[ht]
    \centering
    \includegraphics[width=0.85\linewidth]{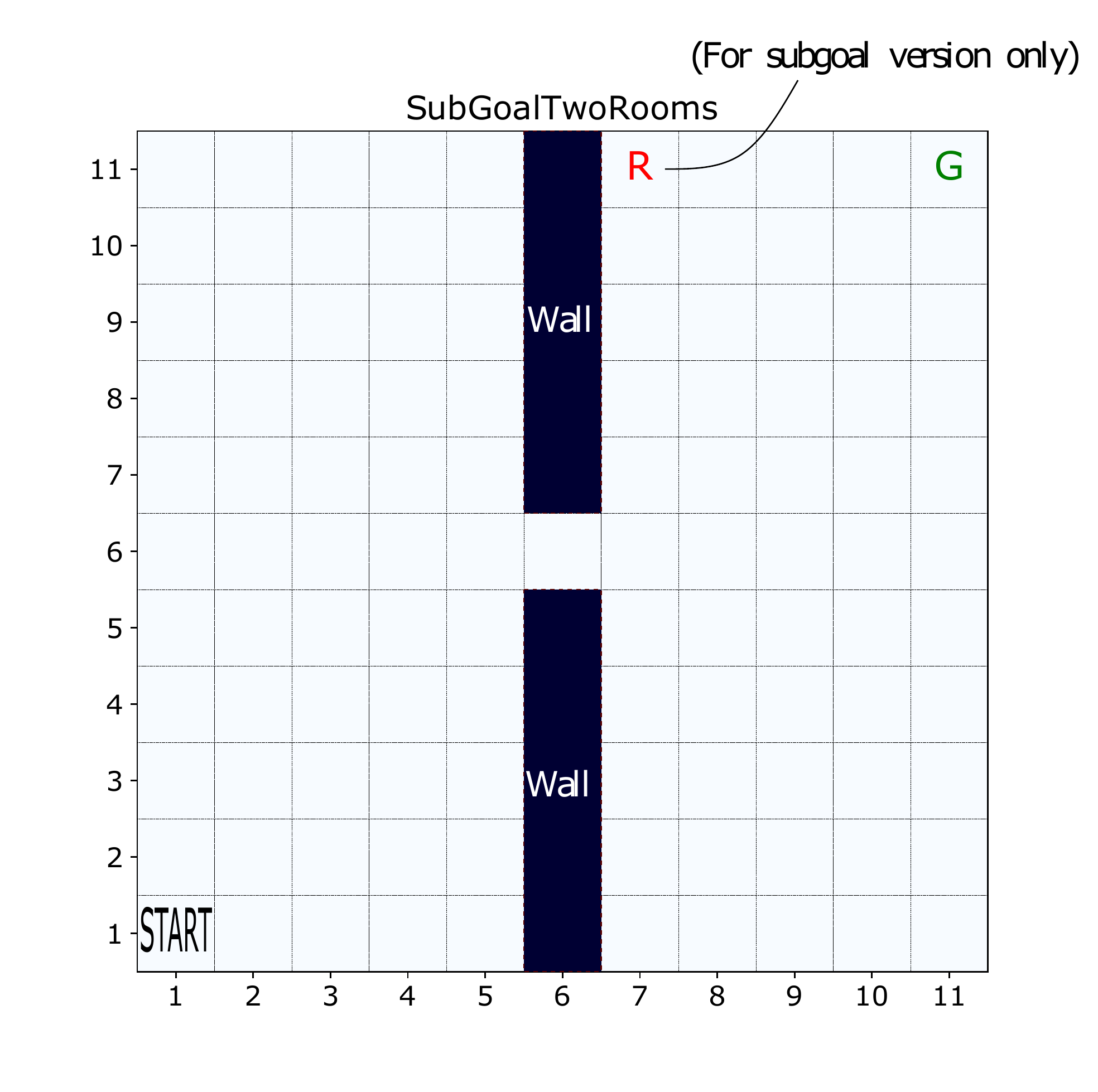}
    \caption{Two Rooms \& SubGoal Two Rooms Environment. The red dot (SubGoal) is represented by R, and the main goal (for both environments) is the green dot shown here as G.}
    \label{fig:env}
\end{figure}


\section{Environments}
Our experiments were tested in the gridworld setting with hand-crafted cumulants. We mainly considered following environments:
\begin{itemize}
    \item \textbf{Two Rooms Environment:} Here, the goal of the agent is to reach the adjacent room by passing through a corridor. As shown in Figure \ref{fig:env}, the agent spawns in the bottom left corner and it has to find its way to the green dot on the top right (the red dot is not relevant in this environment). This seems like an easy task, but for larger dimensions it becomes difficult as the agent has to explore for a long time before it reaches the goal. The agent receives the (x,y) coordinates of its position as its state. The agent gets a terminal reward on reaching the green dot as follows:
    \begin{equation*}
        Reward = 1.0 - 0.9 \times (t/t_{max})
    \end{equation*}
    where $t$ is current time-step and $t_{max}$ is maximum time-step in a episode. Episodes terminates after M=maximum time step of $\text{grid-dim} \times 10$. For this environment, we used only one GVF which gets a cumulant of 1 for crossing the corridor.
    \item \textbf{SubGoal Two Rooms Environment:} This is a modification of the Two Rooms Environment with similar dynamics, except the agent has to collect the red dot and then go to the green dot. The reward for reaching the goal position is as follows 
    \begin{equation*}
    Reward = 
    \begin{cases}
        1.0 - 0.75 \times (t/t_{max}), & \text{if visited red dot} \\
        0.2 - 0.1 \times (t/t_{max}), & \text{else}
    \end{cases}
    \end{equation*}
    The other parameters are exactly the same. For this environment, we append a boolean value of whether the agent has visited the red flag to the (x,y) coordinates to make the states Markov. In this environment, we used one more GVF in addition to the one used in Two Rooms, which receives a cumulant of +1 if the agent visits the red dot.
    \item \textbf{Doorkey Environment:} This is popular environment for benchmarking RL algorithms~\cite{gym_minigrid}, with different tasks. For our experimentations we chose Doorkey environment with $6\times6$ and $8\times8$ grid-size. Board configuration is very similar to Two Rooms environment however, position of corridor and initial position of agent are random. In addition, to enter another room, agent needs to pickup a key, unlock and open the door. Additionally, the actions for going left and right are a bit different than just going left or right. The agent has to first turn to the direction it needs to move in and then go forward. This can be really problematic for EZ-Greedy methods as action repetition would get nowhere and the agent would basically keep on rotating in the same state. Thus, to make the comparison fair, we included options for going left and right for these environments. Additionally, for this environment, we used 2 GVFs. The first GVF gets a cumulant of +1 for picking up the key and the next GVF, gets a cumulant of +1 for unlocking the door.
\end{itemize}
\section{Results}
In this section, we will compare our approach with $\epsilon$-greedy strategy that is widely used as well as the temporally extended $\epsilon$-greedy strategy \cite{DBLP:conf/iclr/DabneyOB21}. The detailed hyper-parameters for reproduction of the results are listed in Appendix B of supplementary material.
\begin{figure}[h]
    \centering
    \includegraphics[width=0.99\linewidth]{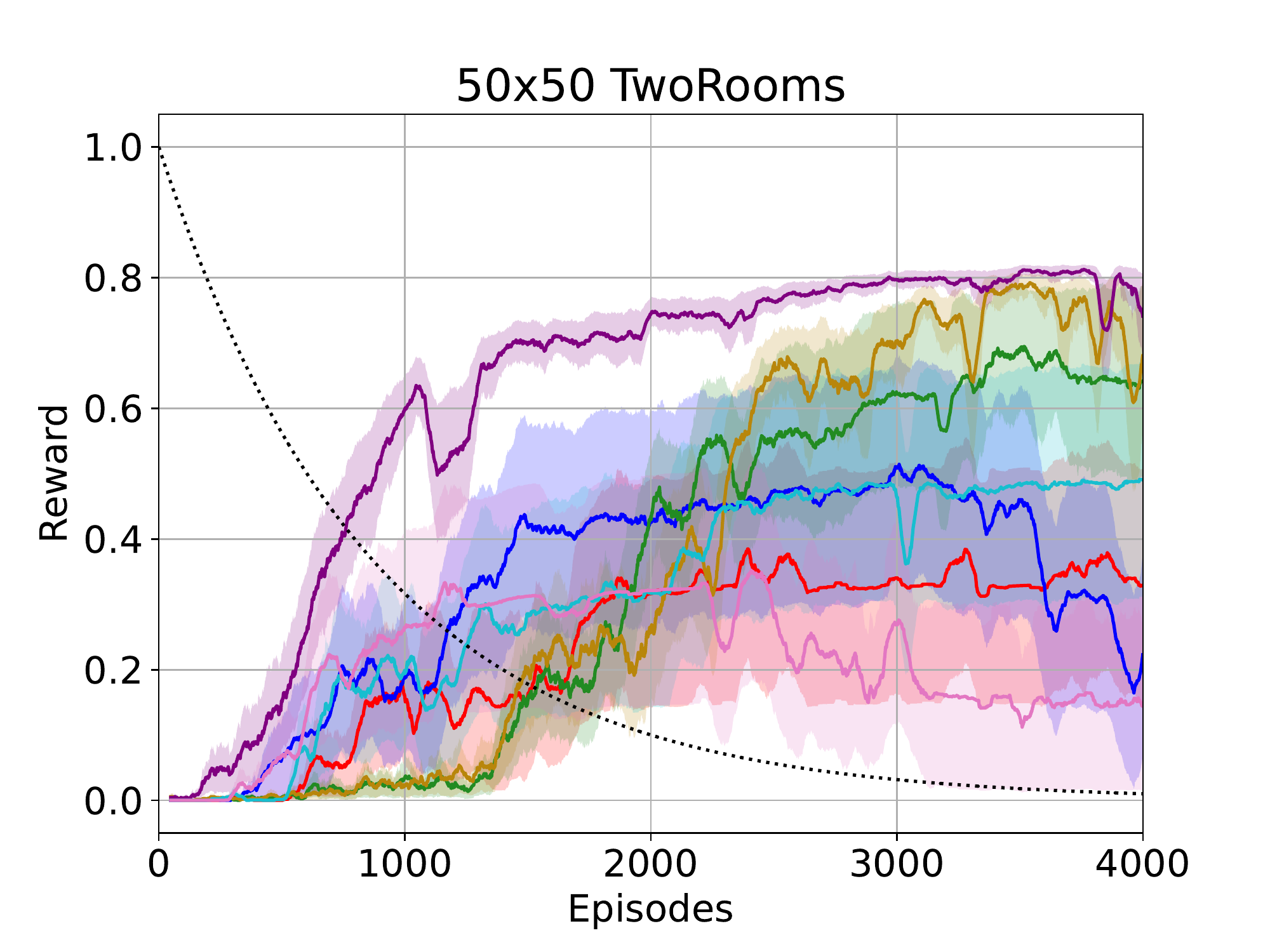}
    \includegraphics[width=0.99\linewidth]{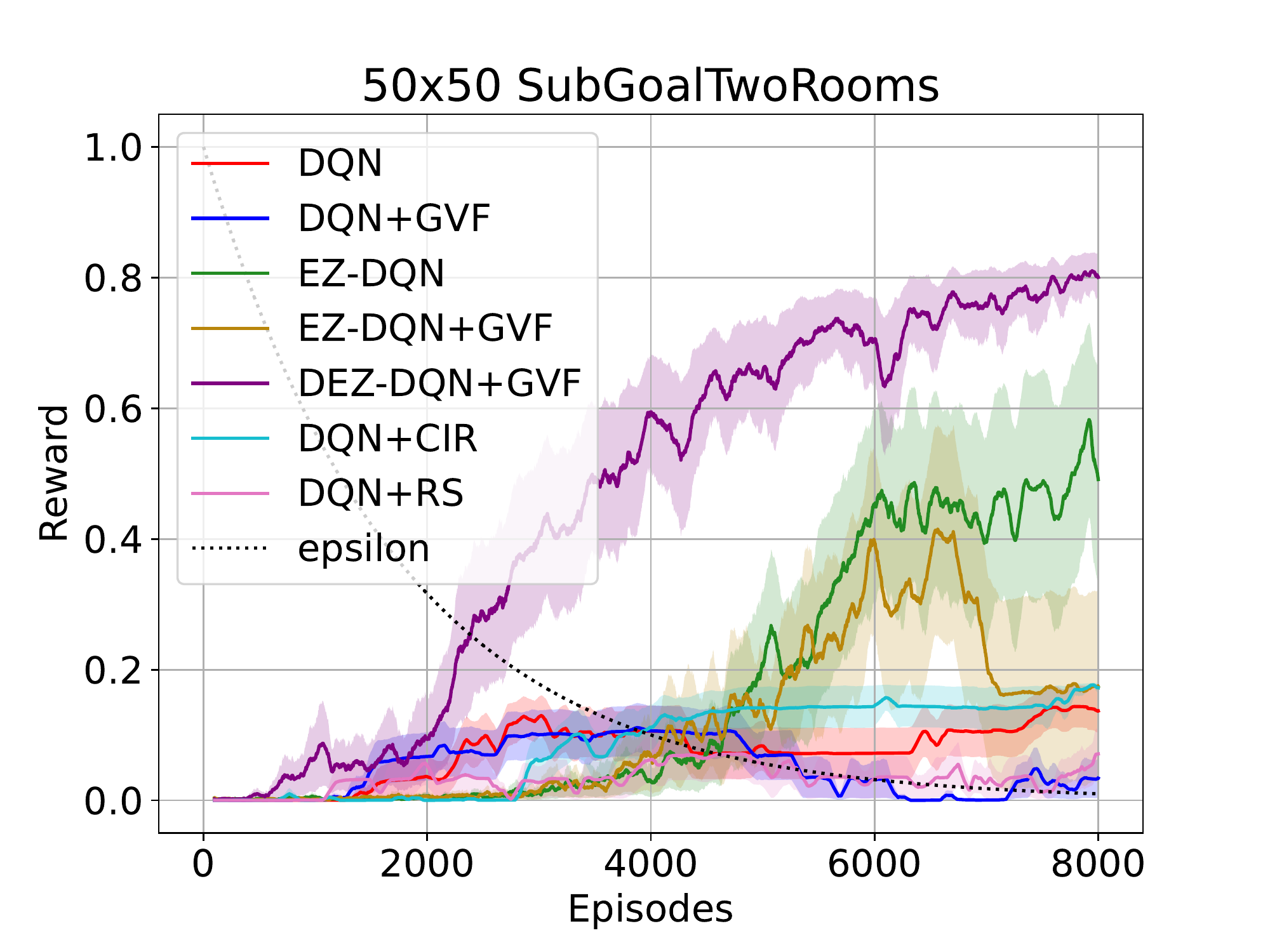}
    \caption{Different Algorithms compared on TwoRooms 50x50 ($Z_{max}=10$) and SubGoal TwoRooms 50x50 ($Z_{max}=30$) Environment, shaded region denotes standard deviation across $5$ seeds.}
    \label{fig:result_toy}
\end{figure}
\paragraph{Algorithms compared:} For our experiments, we compare 7 different algorithms.
\begin{itemize}
    \item The RL algorithm used for our experiments is the DQN algorithm \cite{DBLP:journals/corr/MnihKSGAWR13}, i.e. all the GVFs and the main Q values are learned using DQN. The baseline to compare against would thus be DQN. DQN uses $\epsilon$-greedy exploration with annealing $\epsilon$ and that is what we used for this experiment as well. Details about the variants of DQN used for each of the environments are presented in Appendix B (supplementary).
    \item EZ-DQN is the temporally extended $\epsilon$-greedy version of DQN with EZ-Greedy exploration. The version with repetitive actions is used Two Rooms and SubGoal Two Rooms Environments. But for DoorKey, we used options for going left, right and up instead of repetitive actions. EZ-DQN with repetitive actions for DoorKey is denoted by EZ-DQN(Vanilla).
    \item DQN+GVF is the algorithm that uses GVFs only for representation learning via the common representation which is modified by both the main DQN agent as well as all the GVFs. This still uses $\epsilon$-greedy exploration like DQN.
    \item EZ-DQN+GVF uses the exact architecture as above except the behaviour policy uses EZ-Greedy exploration strategy.
    \item DEZ-DQN+GVF is our algorithm that uses GVFs not only for learning better representations but also for sampling actions for directed exploration
    \item DQN+CIR Count-based Intrinsic Reward policy is count based learning method in which additional intrinsic signal is used to encourage exploration~\cite{STREHL20081309}. The policy is thus trained with the reward composed of two terms $r_t = r_e + \eta \times r_i.$ where $\eta$ is a hyper-parameter and $r_i=1/\sqrt{N(S_t)}$. Authors in \cite{STREHL20081309} propose $1/\sqrt{N(S_t, A_t)}$, we have limited ourselves to counting unique states. This can serve as a proxy for curiosity driven learning.
    \item DQN+RS Reward shaping \cite{laud2004theory} is policy in which positive rewards are given when subgoals are reached, this is an alternative method to incorporate domain knowledge.
\end{itemize}

For our paper, we did not compare any exploratory methods apart from \cite{DBLP:conf/iclr/DabneyOB21}, since this reference is shown to beat other state-of-the-art algorithms. Thus we only used their baseline in addition to epsilon greedy, which is the widely used exploration strategy. We ideally want to ask, what would be a good exploratory strategy that can work well across a bunch of different kinds of environments (both simple and challenging).
\begin{figure*}[h]
    \centering
    \includegraphics[width=0.49\linewidth]{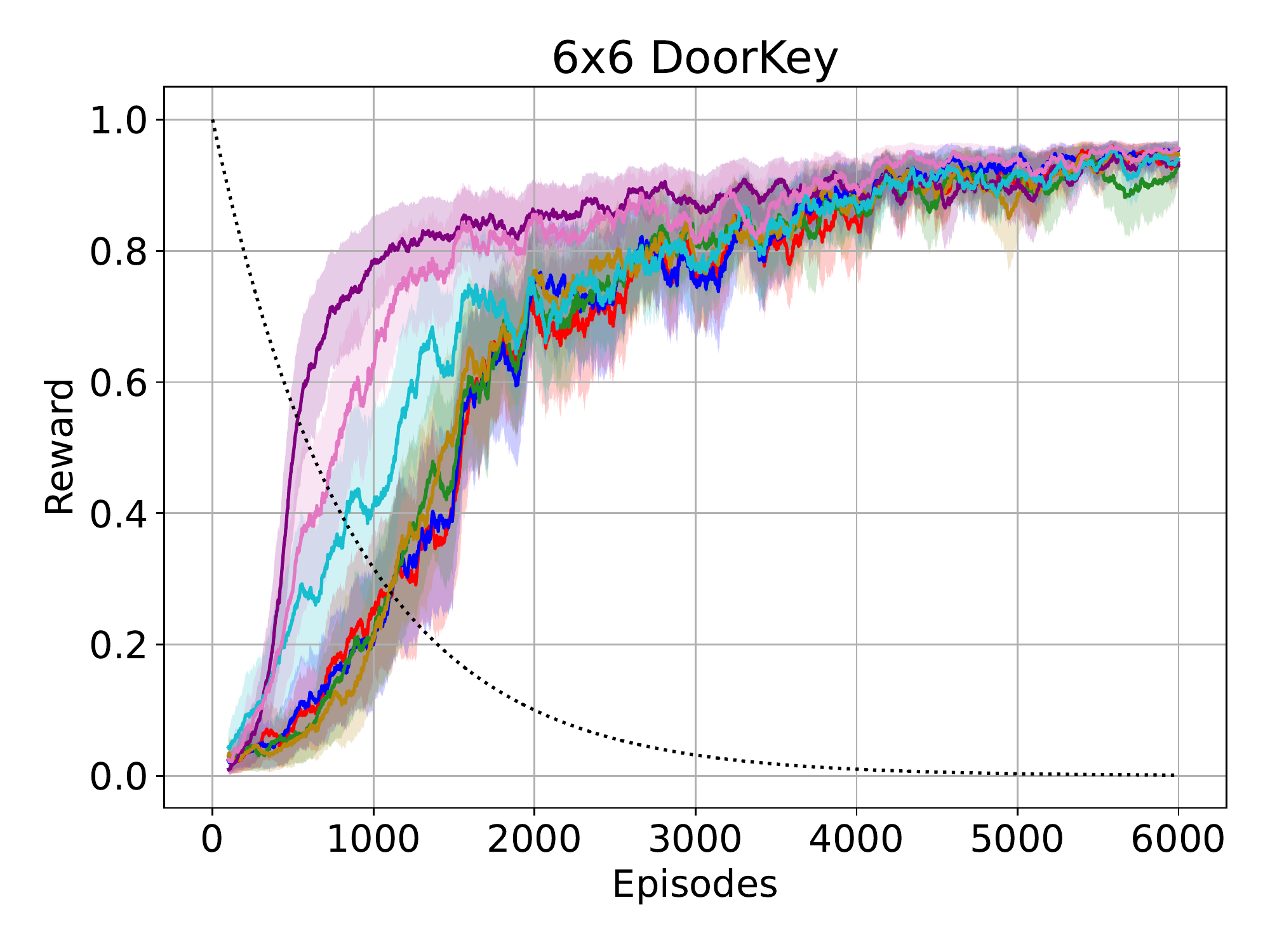}
    \includegraphics[width=0.49\linewidth]{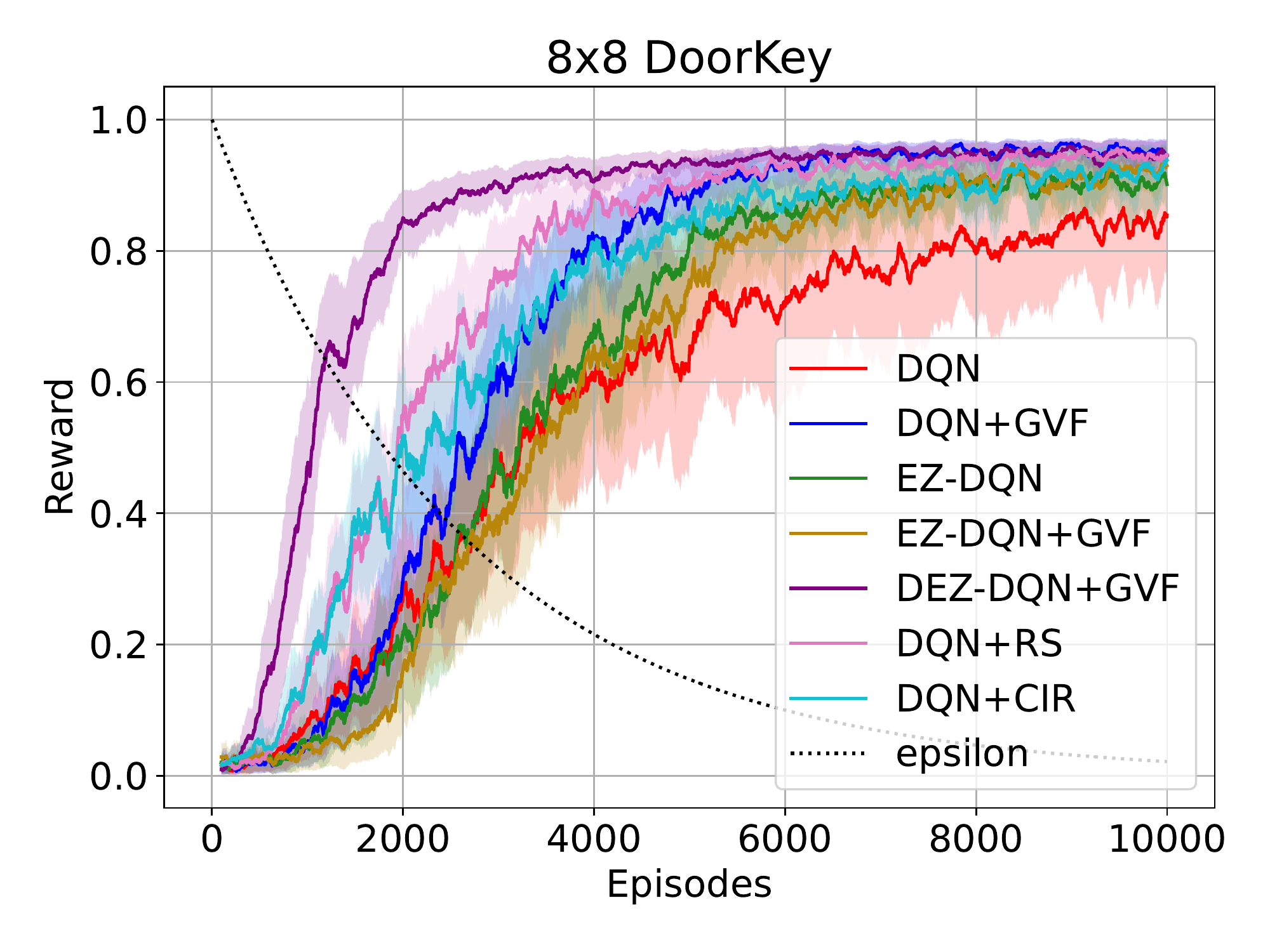}
    \caption{Training plot for DoorKey environment ($Z_{max}=3$), shaded region denotes standard deviation across $5$ seeds.}
    \label{fig:result_doorkey}
\end{figure*}

\subsection{Two Rooms \& SubGoal Two Rooms} The Figure \ref{fig:result_toy} portrays the learning curves of the algorithms across $5$ seeds over $4000$ episodes for TwoRooms and $8000$ episodes for SubGoal TwoRooms environment. The shaded regions denotes the standard deviation across $5$ seeds for all the training curves. 

\begin{figure*}[h]
    \centering
    \includegraphics[width=0.49\linewidth]{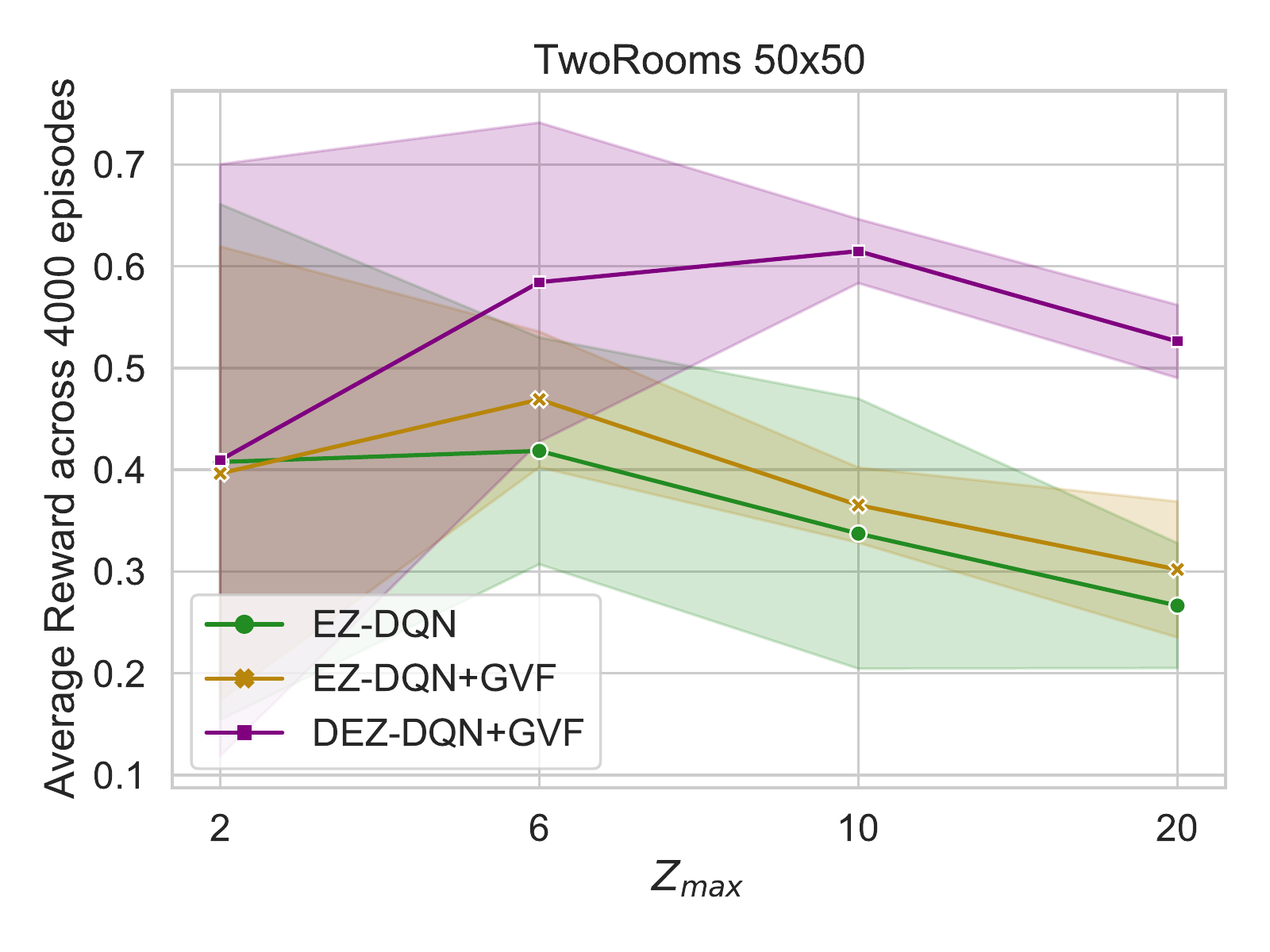}
    \includegraphics[width=0.49\linewidth]{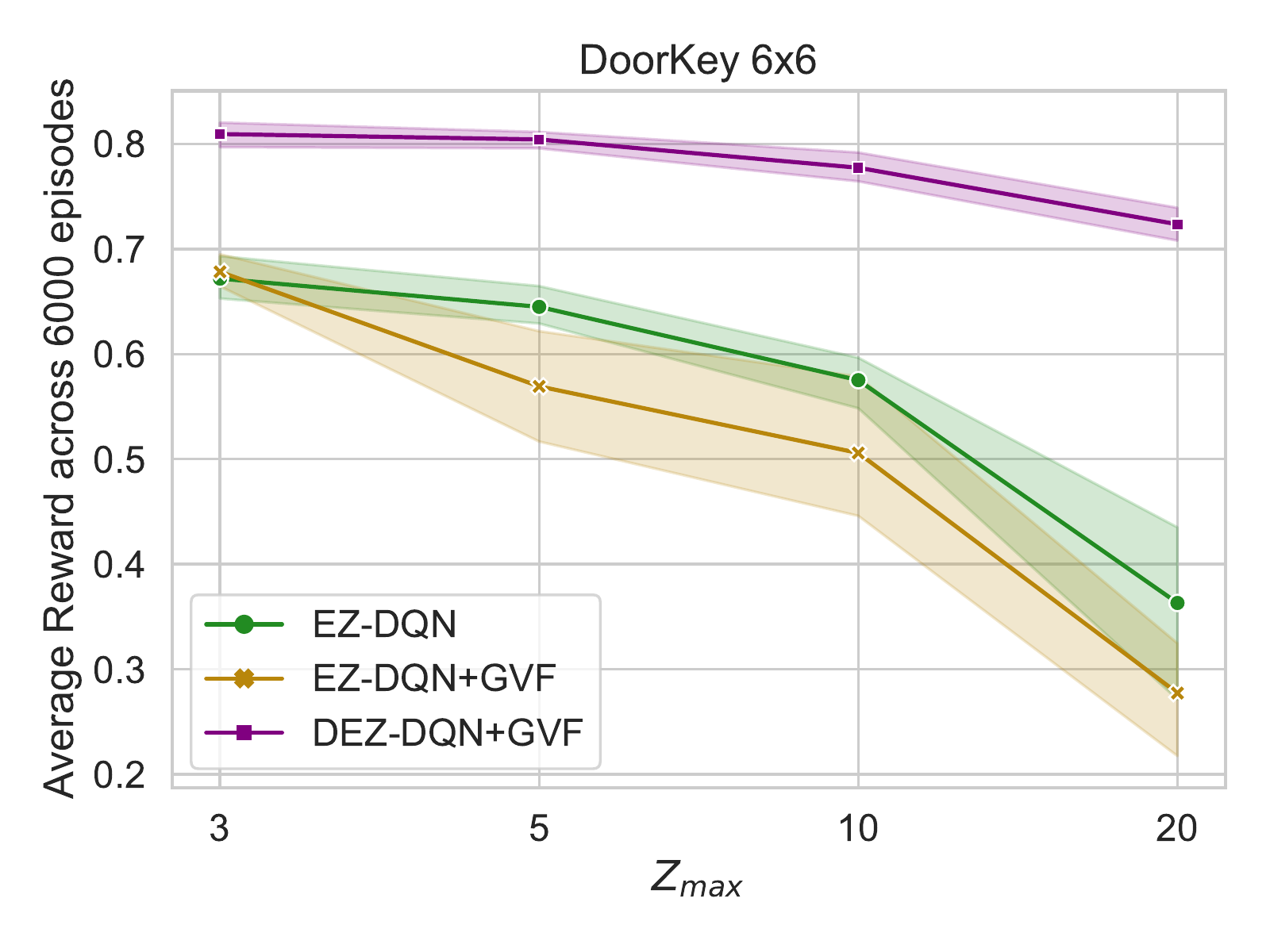}
    \caption{Parameter Sensitivity of EZ-Greedy and DEZ-Greedy strategies with respect to the persistence, $Z_{max}$ in (a) TwoRooms 50x50 and (b) Doorkey 6x6 Environment. Shaded region denotes 95\% confidence interval across 5 random seeds.}  
    \label{fig:persistence}
\end{figure*}

\noindent It is evident that EZ-Greedy and in particular taking repeated actions helps in exploring the state space and there is a marked difference between algorithms that use persistence and normal $\epsilon$-greedy.  More so in SubGoal Two Rooms environment where $\epsilon$-greedy algorithms have converged to sub-optimal values. Adding directed exploration with the use of GVFs makes the RL agent outperform EZ-Greedy because of the exploration of useful state spaces as the GVFs are designed to stimulate goal-driven exploratory behaviour. Another interesting thing worth noting is that even adding GVFs without using any exploratory strategy (DQN+GVF blue) helps in better performance in Two Rooms environment compared to its counterpart (DQN red). This is likely due to GVFs learning useful representations which can improve performance. However, there is performance degradation in SubGoal Two Rooms and we believe this is due to policies converging to sub optimal values and inherent sample inefficiency. A similar pattern is observed for EZ-Greedy with GVFs there is an improvement in performance in the Two Rooms environment, while slight drop in SubGoal Two rooms environment.  In addition, DQN+CIR is able to outperform vanilla DQN due to better exploration characteristics in Two Rooms however, the count-based exploration strategy is still not sufficient to elevate the DQN from suboptimal point in SubGoal Two rooms environment.  

\begin{figure*}
    \centering
    \includegraphics[width=0.49\linewidth]{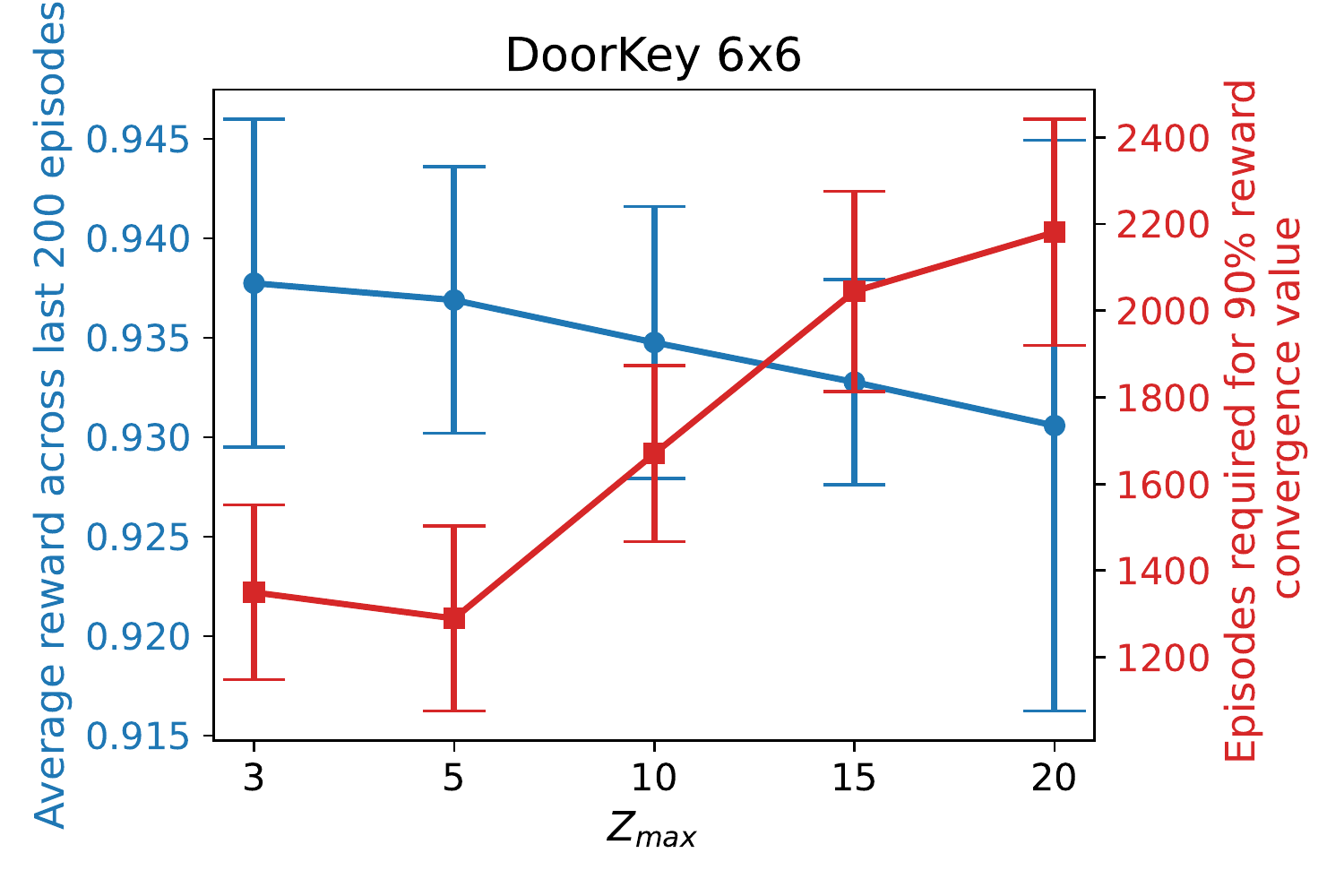}
    \includegraphics[width=0.49\linewidth]{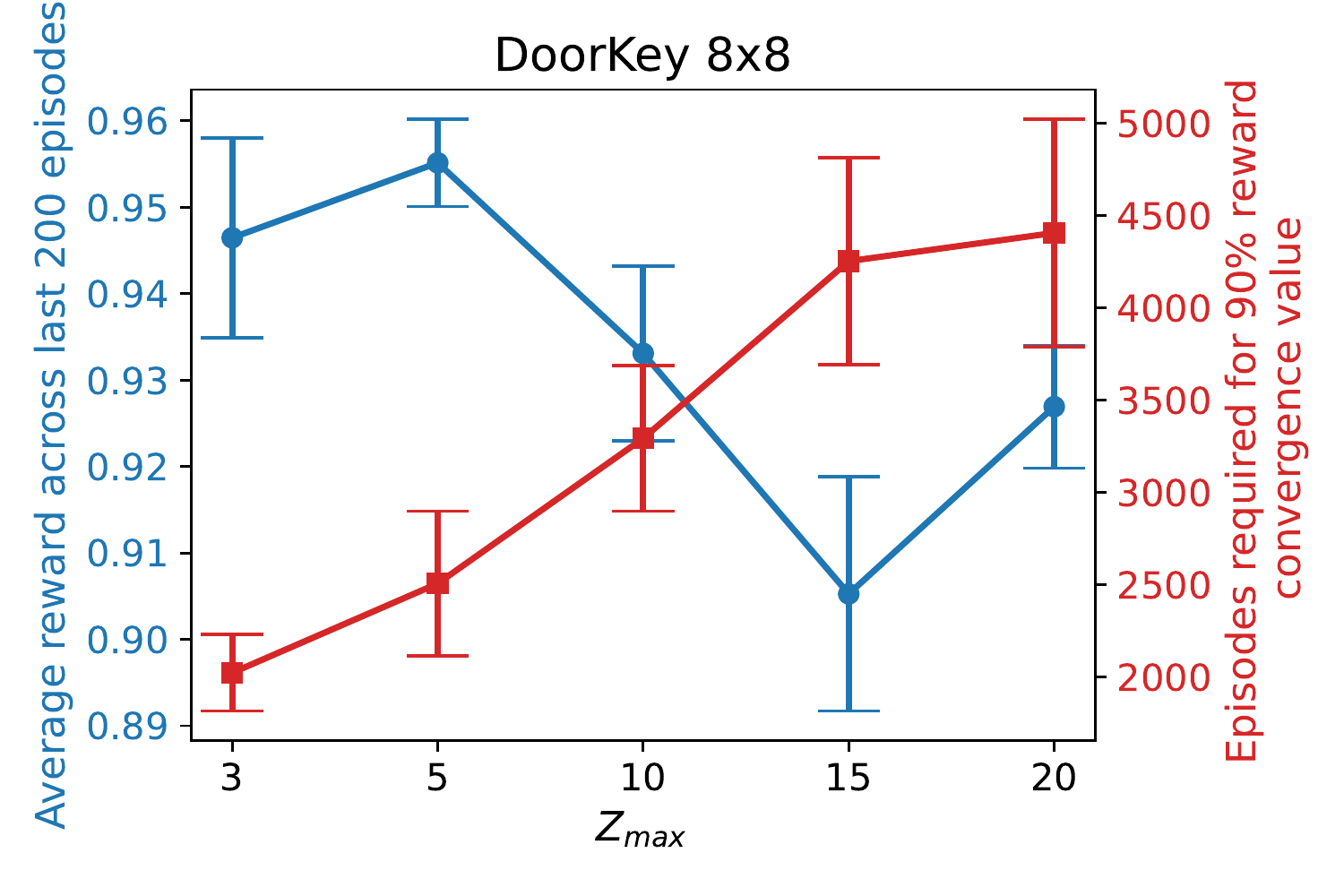}
    \caption{DEZ-Greedy performance for different persistence values on Doorkey environment. The Y-axis on the left represents the average reward across the last $200$ games while the axis on the left displays the number of episodes DEZ-DQN+GVF took to reach $90$\% of the final convergence value of the reward. Error bar denotes the standard deviation across $5$ random seeds.}
    \label{fig:doorkey_zmax}
\end{figure*}

\subsection{DoorKey}
In Figure~\ref{fig:result_doorkey} we present training results for 2 grid sizes; $6 \times 6$  and $8 \times 8$  of the Doorkey environment from Gym MiniGrid~\cite{gym_minigrid}. For both the gridsize we can observe DEZ-DQN+GVF converging early compared to other algorithms, the difference is more pronounced in higher grid size. For $8 \times 8$ size, we see higher variance across random seeds for all the baselines whereas DEZ-DQN+GVF has relatively low variance and shows a similar trend to $6 \times 6$ providing evidence of robustness across grid sizes, as observed earlier. It is worth noting that having reward shaping (DQN+RS) which is a proxy for introducing additional information into the learning does help the vanilla DQN and shows improvement over other baselines. Similarly, DQN+CIR does get a steep rise in rewards thanks to reaching novel states earlier than other baselines however it does taper off as the training progress. 

It is also interesting to note there is little or no improvement in the performance of EZ-DQN+GVF over EZ-DQN for $8 \times 8$ DoorKey which is in stark contrast to Two Rooms and SubGoal Two Rooms environments. We believe this is due to the off-policy divergence that can happen when the greedy policy with respect to each of the GVFs is very different from the behaviour policy of the main RL agent (also amplified by k repeated actions). This phenomenon discourages the use of GVFs for representation learning particularly for Deep RL methods because the divergence in GVFs values results in degradation in performance. However, our method periodically takes actions from the GVFs, it helps in avoiding diverging values.

\begin{figure}[h]
    \centering
    \includegraphics[width=0.98\linewidth]{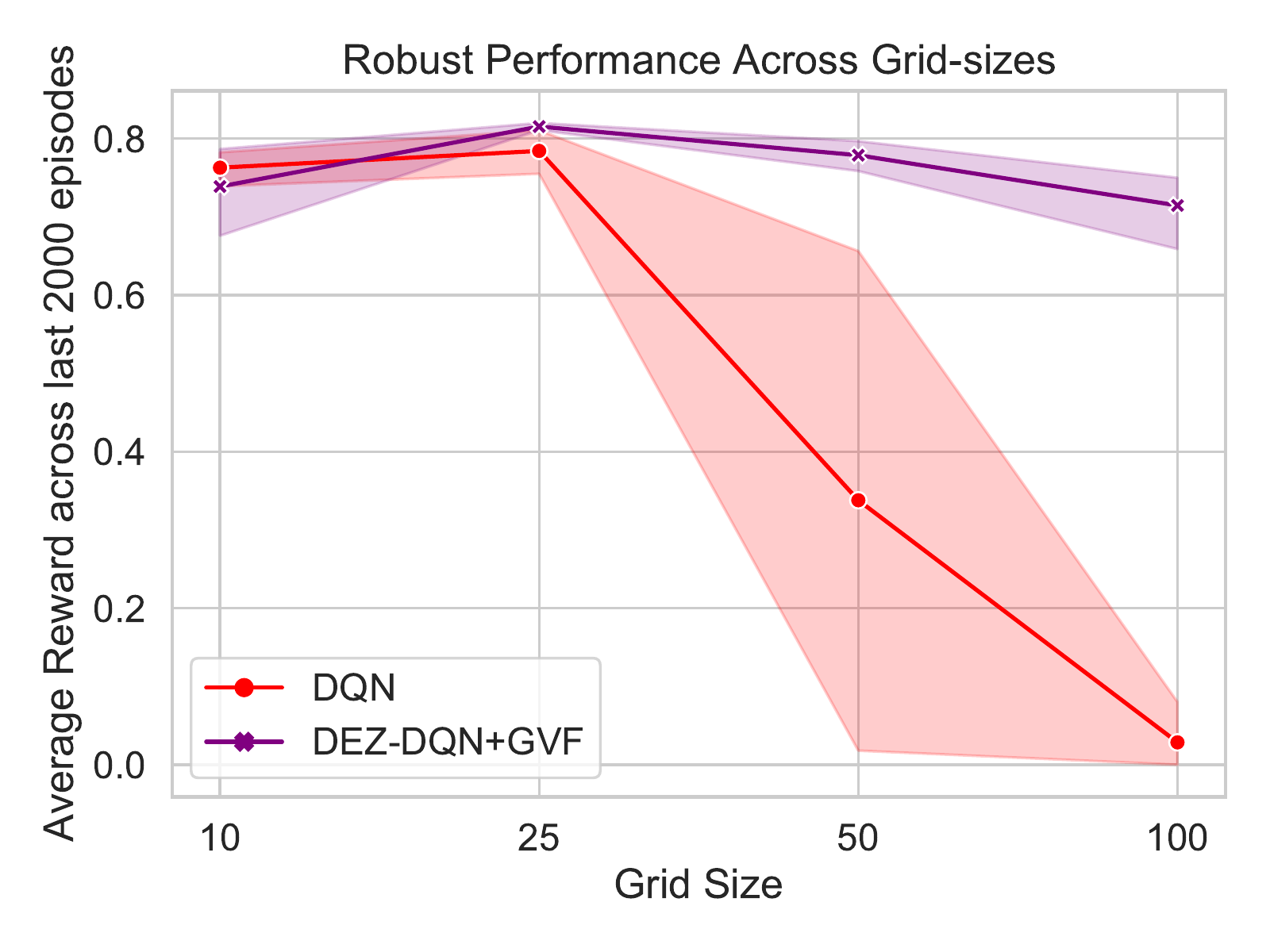}
    \caption{Performance comparison of DQN and DEZ-DQN+GVF for various grid sizes of TwoRooms Environment}
    \label{fig:subgoal_grid}
\end{figure}


\paragraph{Sensitivity with respect to Persistence $Z_{max}$:}
The performance of the algorithm will be determined by the persistence values chosen. For example, a persistence value of 1 for EZ exploration reduces to $\epsilon$ greedy exploration. In Figure \ref{fig:persistence}, we plot the sensitivity of the algorithms across different persistence values for a) $50 \times 50$ Two Rooms environment and b) door key environment with $6 \times 6$ grid size. We can observe that there is an optimal value of persistence which provides the best performance across the algorithms, while lower and higher values result in degradation of performance. Lower values would essentially lead to $\epsilon$ greedy exploration whereas high values would result in agent sticking to boundaries in case of random sampling. However, DEZ is much more robust to these kinds of situations and provides much more reliable performance.

\noindent To further understand the effect of $Z_{max}$ on convergence characteristics, Figure \ref{fig:doorkey_zmax} portrays the performance of DEZ-DQN+GVF on $6 \times 6$ and $8 \times 8$ DoorKey with respect to increasing $Z_{max}$. Two Y-axis represent the average reward across the last $200$ episodes (blue) and the number of episodes taken to reach the 90\% of convergence value of the reward (red). The spread of values across the left Y-axis (blue) is very minimal for $6 \times 6$, and within a reasonable range for $8 \times 8$. As expected only change noticeable change happens in the red line, indicating higher $Z_{max}$ values take longer to reach convergence value due to the inherent smaller grid size of the environment. With a high value for $Z_{max}$ in a smaller grid size EZ-DQN algorithm leads the agent to the edge of the grid sizes, whereas for DEZ-DQN+GVFit quickly achieves the GVF goal. In both of these algorithms actions are wasted and hence slower convergence with higher $Z_{max}$ however, with DEZ-DQN+GVFfinal performance is still not degraded significantly as the agent is closer to achieving the goal than before. 

\paragraph{Sensitivity with respect to grid size:}
One aspect that we wanted to target is better performance for both simple and hard domains. In Figure \ref{fig:subgoal_grid} we plot the mean average reward for $\epsilon$-greedy and DEZ-Greedy method. With smaller grid size, $\epsilon$-greedy does really well but it fails to work for higher grid sizes. However, DEZ-greedy not only works for the higher grid sizes, but also has comparable performance to $\epsilon$-greedy in smaller domains as well. This is an interesting result because even it shows how this strategy is robust to the hardness of the environment with respect to exploration. Thus DEZ-Greedy can be used universally for all types of environments. Here for DEZ-Greedy method for grid-sizes 10, 25, 50 and 100 we use 5, 5, 10 and 20 $Z_{max}$ values respectively. 
\begin{figure}[h]
    \centering
    \includegraphics[width=0.95\linewidth]{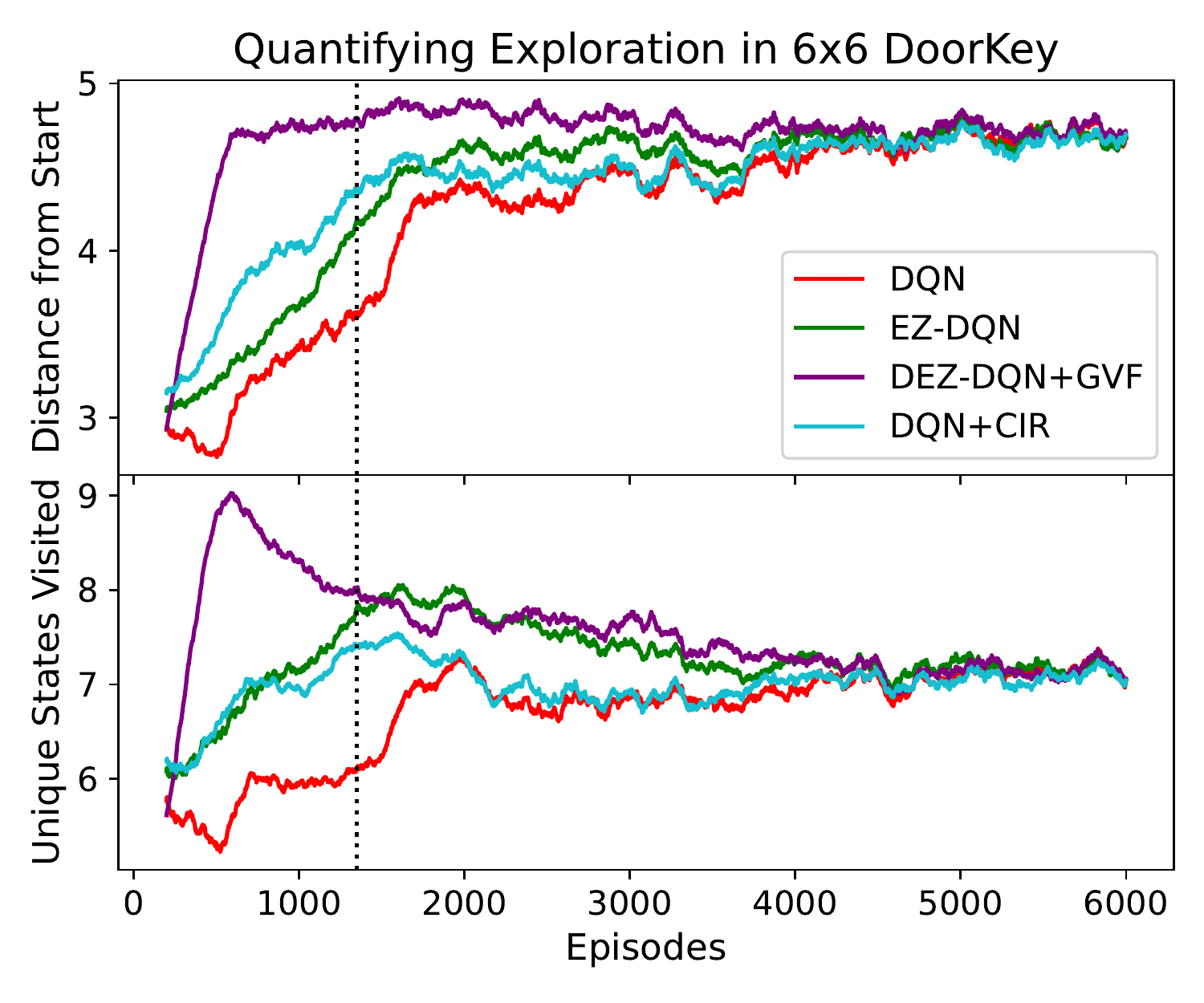}
    \includegraphics[width=0.99\linewidth]{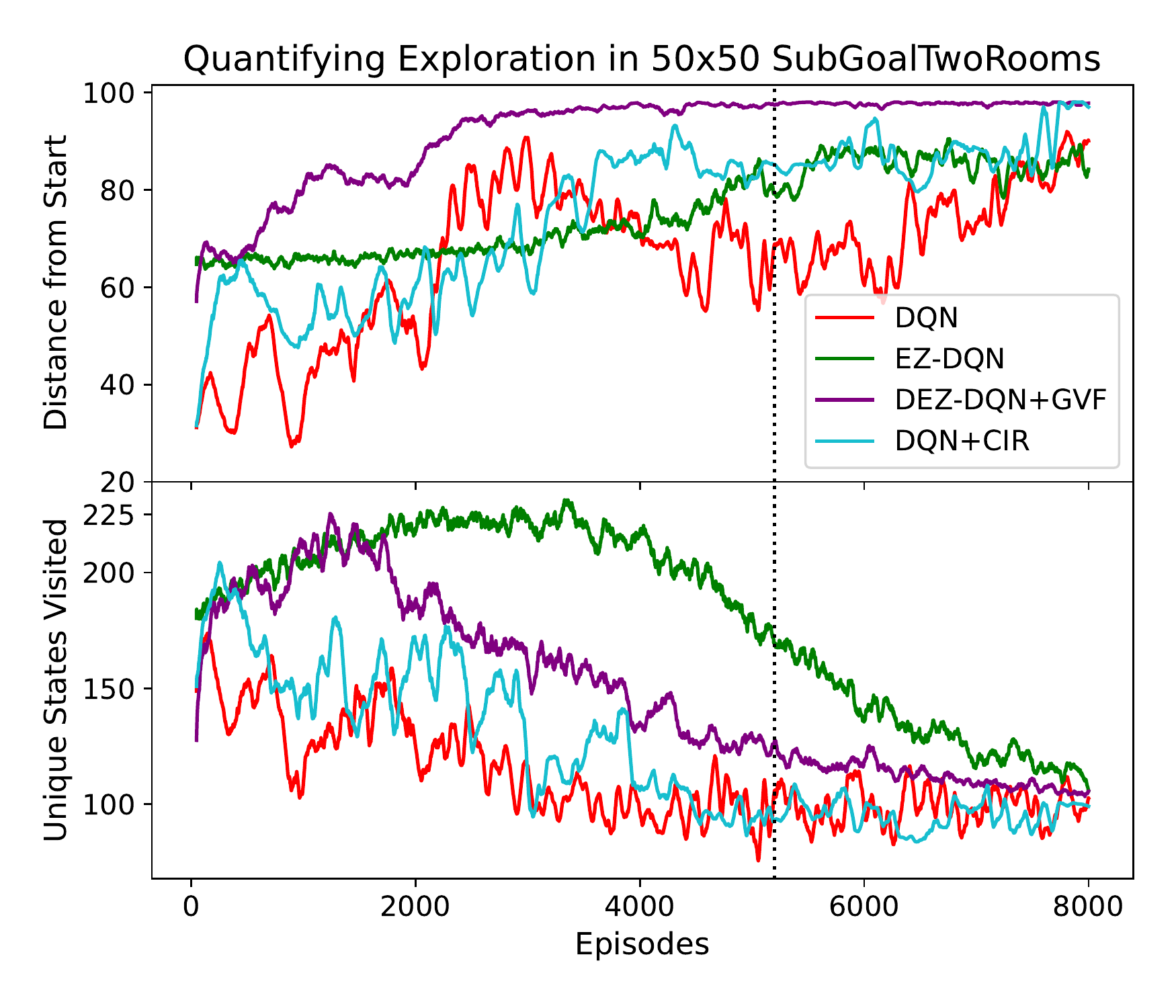}
    \caption{This plot tries to quantify exploration by comparing distance and unique states visited by normal epsilon greedy and directed epsilon greedy method. The graph is plotted with aggregating values for 100 episodes. a) 6x6 DoorKey b) 50x50 SubGoal TwoRooms}
    \label{fig:exploration}
\end{figure}
\begin{figure}[h]
    \centering
    \includegraphics[width=0.97\linewidth]{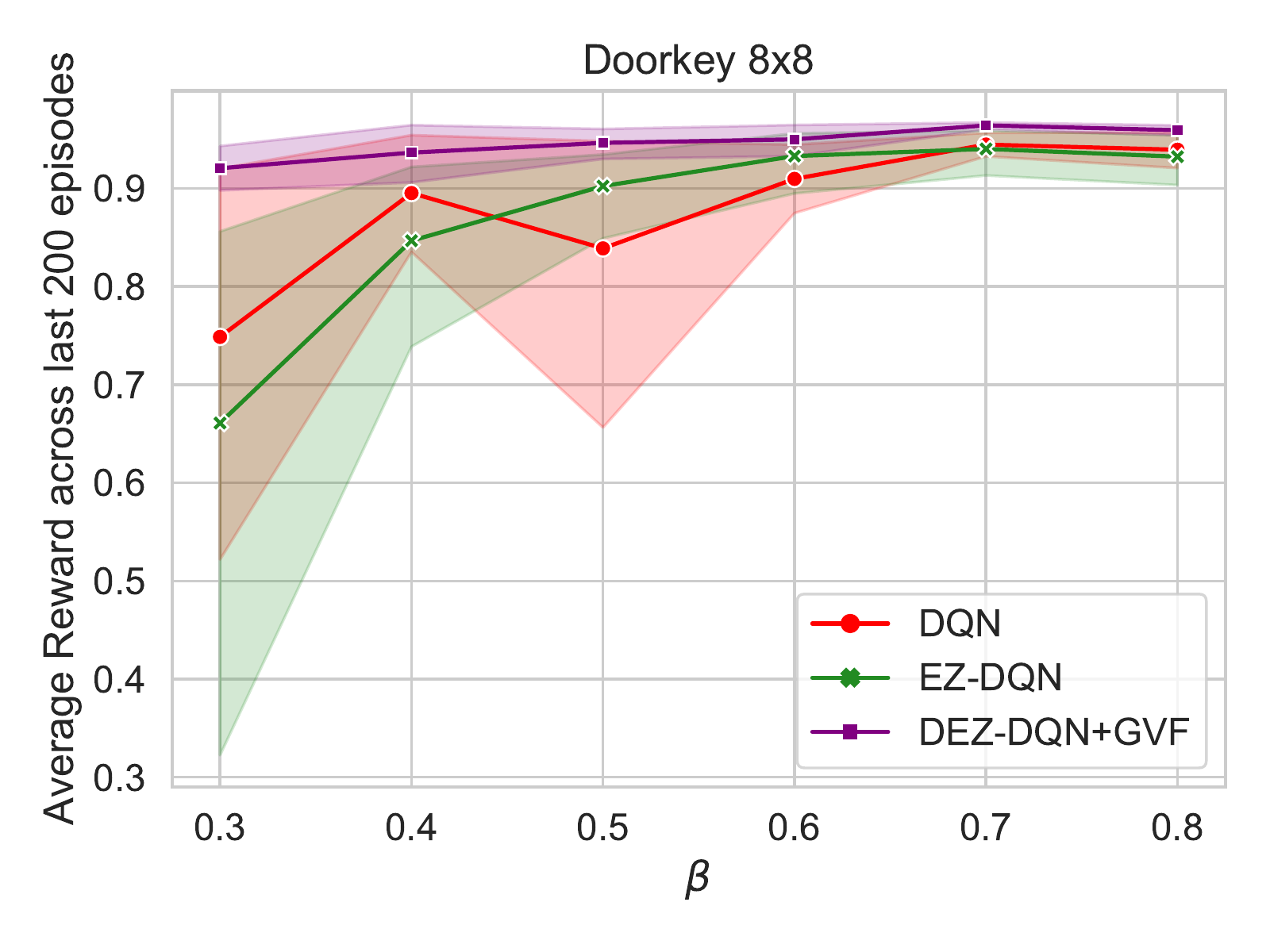}
    \includegraphics[width=0.97\linewidth]{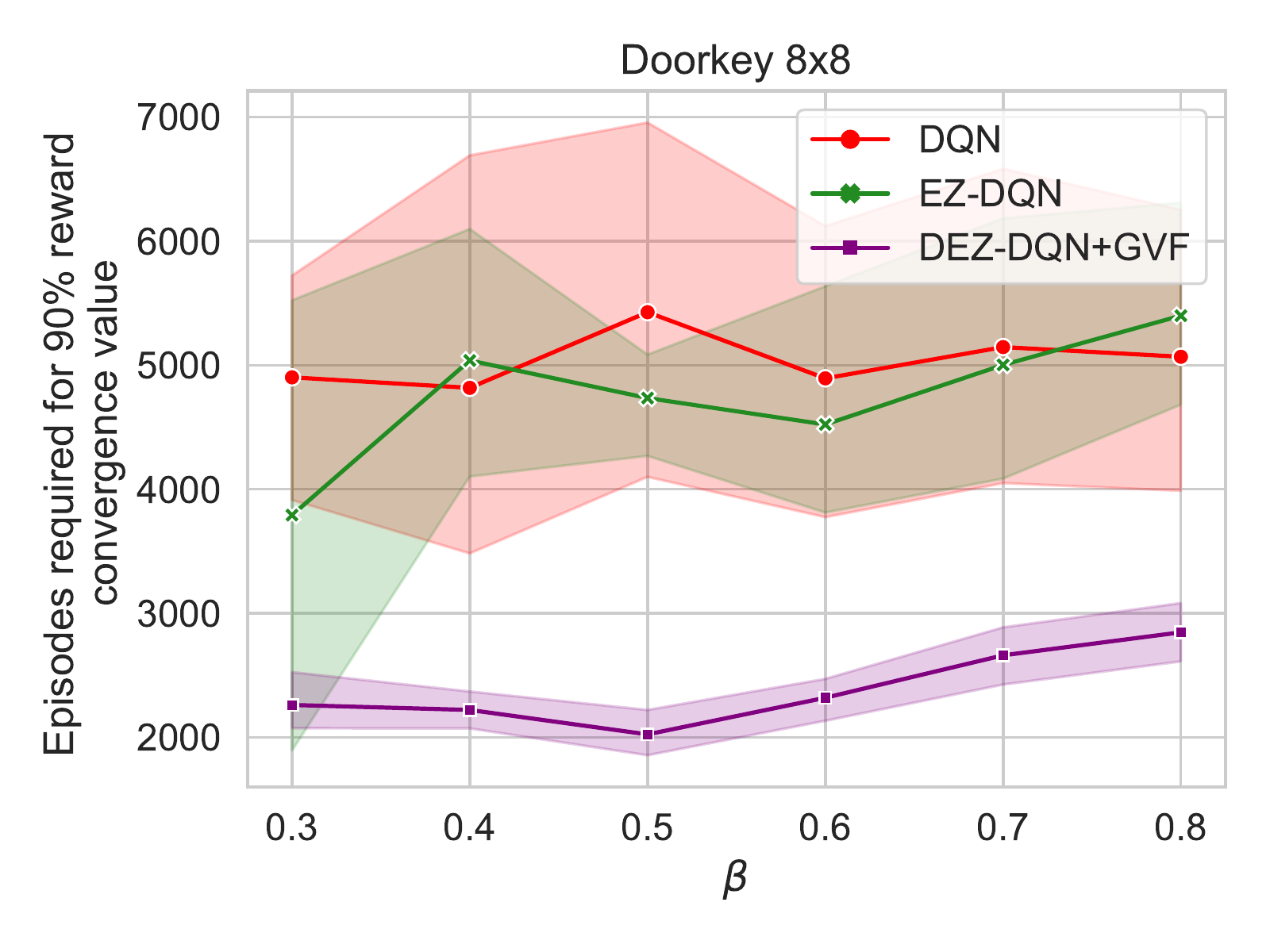}
    \caption{Performance of different algorithm for different $\beta$ values. This plot tries to quantify the robustness of different algorithm with respect to rate of epsilon decay for exploration. Shaded region denotes 95\% confidence interval across 5 random seeds}
    \label{fig:doorkey_beta}
\end{figure}

\paragraph{Quantifying Exploration:} 
Having a higher reward is one way of ascertaining that the agent has explored important parts of the state space faster and as a result is able to converge much faster. However, a direct comparison of the exploration strategies can be made by using  visitation frequency of unique states. Figure \ref{fig:exploration} presents a comparison of various exploration strategies directly by plotting the farthest distance from the initial state and the number of unique states visited. Distance from the initial state is also quantification of exploration for this environment as the distance would increase when the agent can unlock the door and visit the next room. Vertical line indicates the episode when DEZ-DQN+GVF reached 90\% of its convergent value. We can observe that directional exploration is helping DEZ to not only explore farther from the initial state but also reach unique states quickly. In addition, number of unique states quickly goes down as it reaches it 90\% convergence value providing enough evidence that it has found its optimal policy way early compared to baselines.

To further understand the robustness of DEZ with respect to exploration, we define epsilon decay as
\begin{align*}
\epsilon_{decay}=(0.01)^{\frac{1}{episodes \times \beta}}
\end{align*}
where $\beta \in (0,1]$ is a hyper-parameter. So for small $\beta$ value the rate of epsilon decay will be fast and the agent will have less time for exploration. Figure~\ref{fig:doorkey_beta} a) shows the reward achieved while training with different $\beta$ epsilon schedule, we can observe that DEZ-DQN+GVF has little to no variance even during less value of $\beta$ and minute drop in convergence value. This suggests that it is much more robust to the exploration schedule and achieves better results even with less exploration. This hypothesis is further reinforced by results in Figure~\ref{fig:doorkey_beta} b), where we can observe that number of episodes required to reach 90\% of the convergence value is much lower compared to baselines and increases marginally with respect to $\beta$. 

\paragraph{Algorithm Overhead:}
 In general, online Reinforcement Learning does not suffer from the problem of computational complexity, but more from sample inefficiency. Specially in real world settings when the number of interactions are limited, DEZ-Greedy can be really useful, because it requires much fewer samples for better performance. In terms of compute, yes, it does add compute because it involves training General Value Functions, in addition to the main RL network. However we note that this is parallelizable during training; and it makes no effect on post-training inference since we use only the main policy. Training time of DEZ and EZ for 50x50 two-rooms env for 4000 episodes is 1hr 7min and 55 min respectively. The hardware used for the experiments consist of an Nvidia A100 GPU with 14 vCPUs.

\section{Conclusion and Future Work}
Exploration is one of the critical aspects for good performance for Reinforcement Learning, especially in games. EZ-Greedy \cite{DBLP:conf/iclr/DabneyOB21} has been shown to perform better overall in a variety of Atari games compared to curiosity driven methods which specialize in exploration for some specific games. DEZ-Greedy incorporates aspects of both EZ-Greedy and auxiliary task learning along with learning richer representations and is more robust as well. Thus, it should work well in a variety of games. From the experiments, DEZ-Greedy seems to learn much faster because of a better exploration strategy along with improved representation learning. Intuitively as well, this makes sense as actions generated from the GVFs take the agent into states which are useful for optimizing towards the main goal. Additionally, this architecture is very flexible as we only require a scalar reward to learn the GVFs.

The next steps would be to try to adapt it to environments which can learn the cumulants automatically from the environments \cite{DBLP:conf/nips/VeeriahHXRLOHSS19}. It can also be used for transfer learning in procedurally generated environments as the GVFs learned can help in generating useful directed exploration for newer tasks as well.


\bibliographystyle{ACM-Reference-Format} 
\bibliography{sample}


\appendix

\section{}

\begin{figure}[ht]
	\centering
	\includegraphics[width=0.98\linewidth]{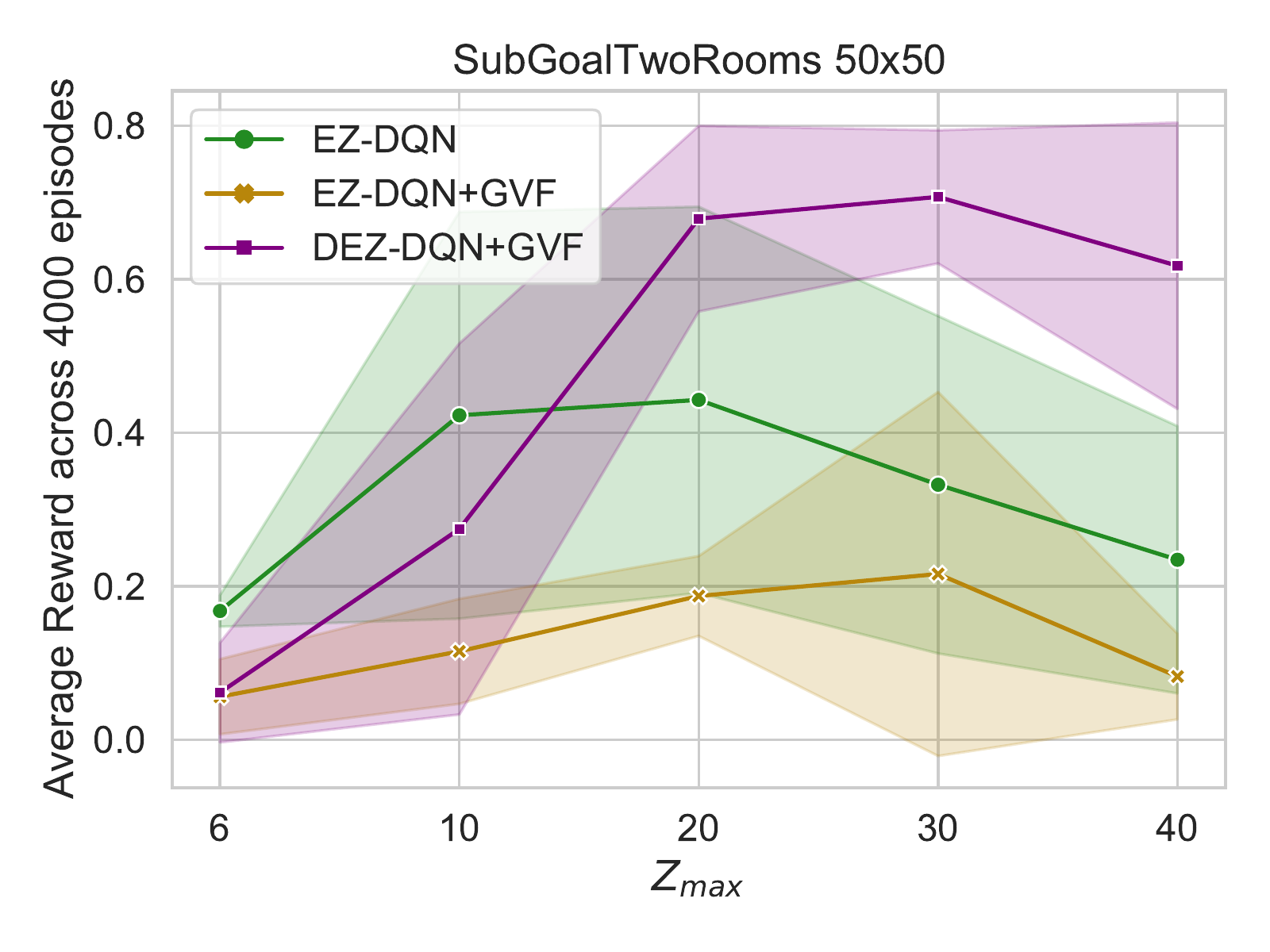}
	\caption{Parameter Sensitivity of EZ-greedy and DEZ-greedy strategies with respect to the persistence, $Z_{max}$ in SubGoalTwoRooms 50x50. Shaded region denotes 95\% confidence interval across 5 random seeds}  
	\label{fig:persistence_subgoal50}
\end{figure}

In Figure \ref{fig:persistence_subgoal50}, we plot the sensitivity of the algorithms across different persistence values for sub-goal two rooms environment with $50 \times 50$ grid size. For lower values of persistence we observe that all the the algorithms have sub-optimal performance. As we go in increasing the persistence values the performance of DEZ-greedy method increases whereas other methods are unable to perform similarly. We can see that having higher persistence values leads to better rewards, thus DEZ-greedy method is an better option to proceed with environments having higher grid sizes.\\

\begin{figure}[ht]
	\centering
	\includegraphics[width=0.98\linewidth]{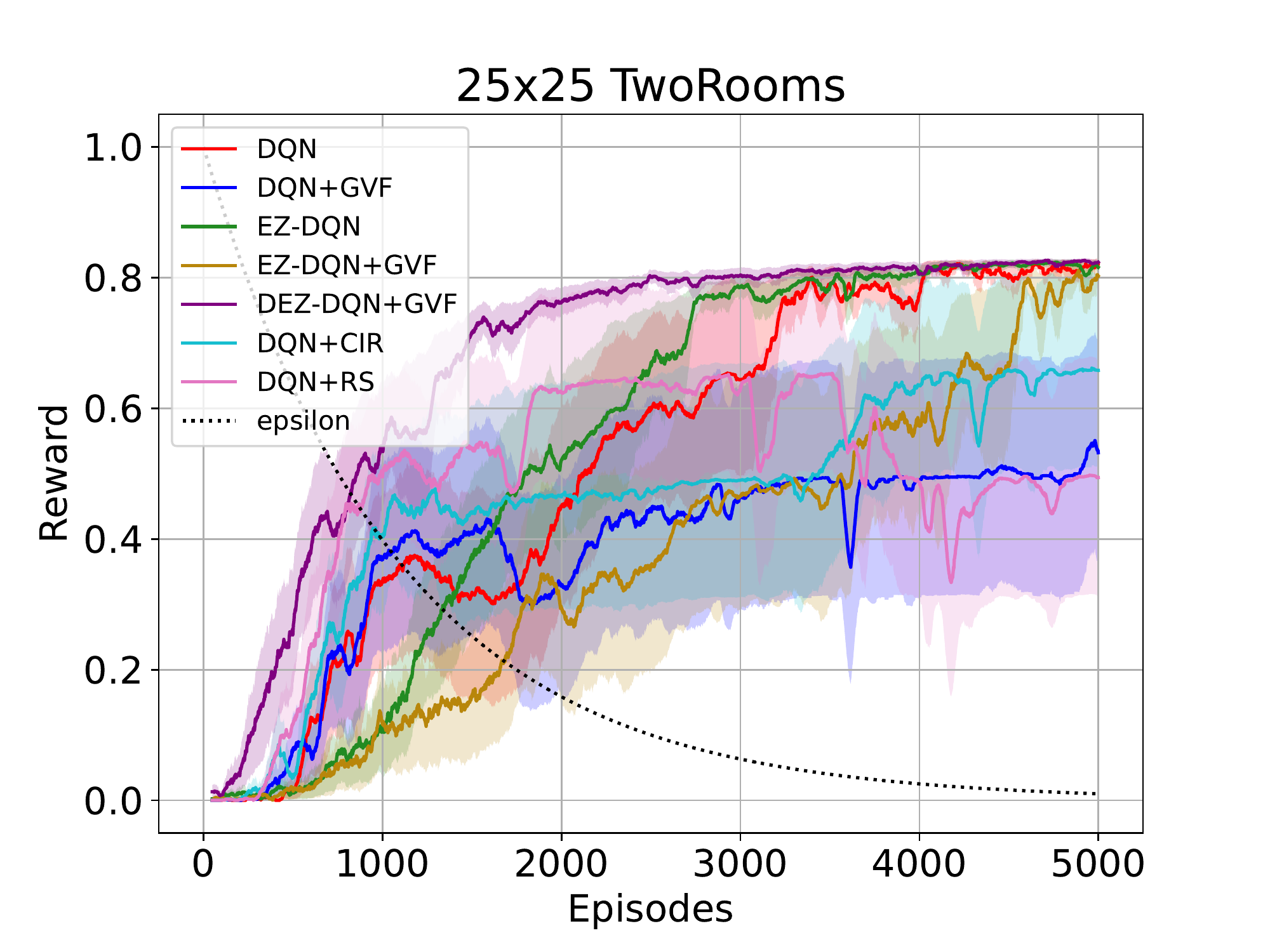}
	\caption{Different Algorithms compared on TwoRooms 25x25 ($Z_{max}=5$) Environment, Shaded region denotes standard deviation across $5$ seeds}
	\label{fig:tworooms_25x25_reward}
\end{figure}

In Figure \ref{fig:tworooms_25x25_reward} we plot the learning curve of the algorithms across $5$ runs over $5000$ episodes for $25 \times 25$ two rooms environments. Here also it is evident that DEZ-greedy performs better compare to the baseline algorithms. The results are same as the $50 \times 50$ two rooms environments results discussed in the main text.\\

\begin{figure*}[ht]
	\centering
	\includegraphics[width=0.5\linewidth]{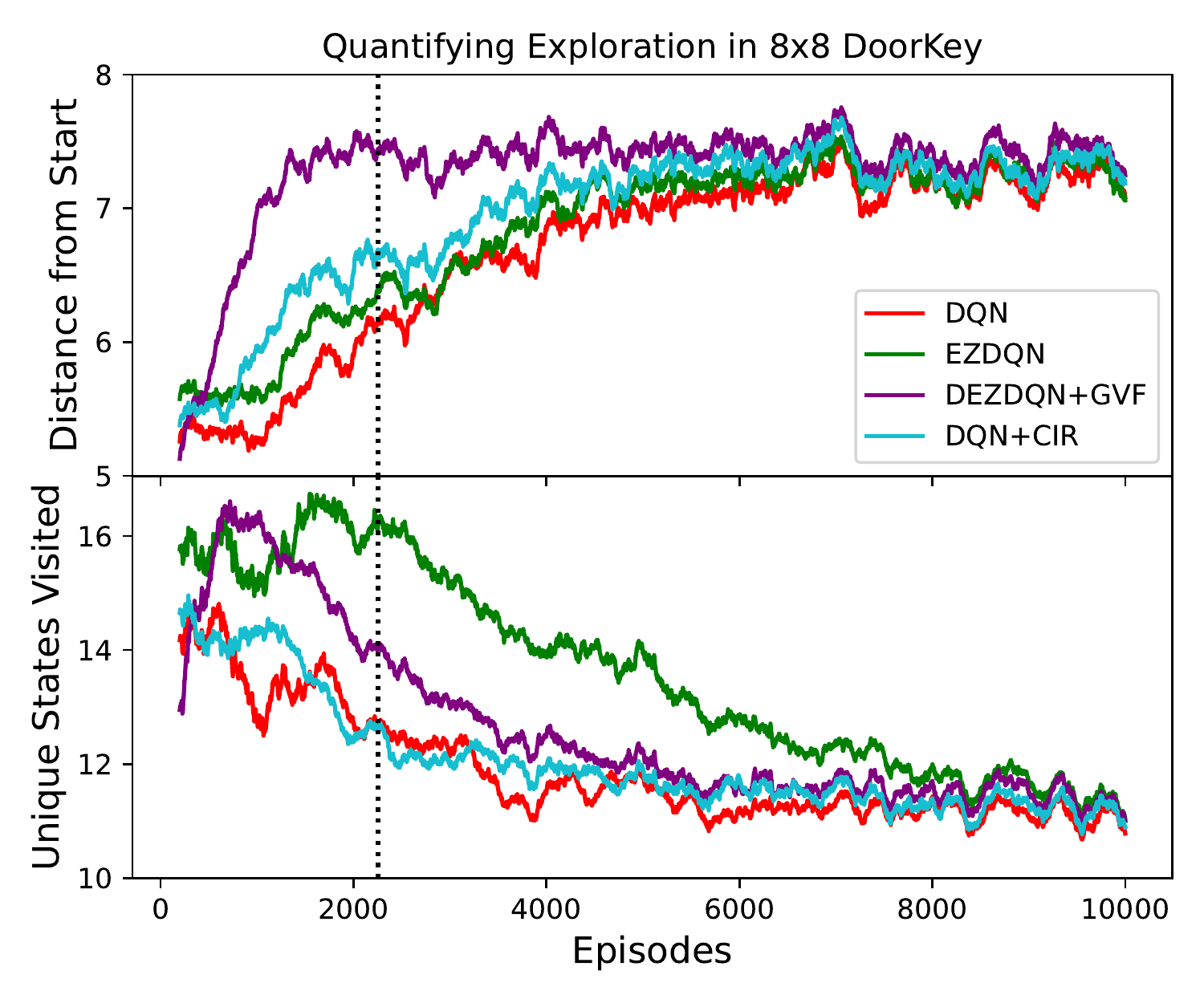}
	\includegraphics[width=0.5\linewidth]{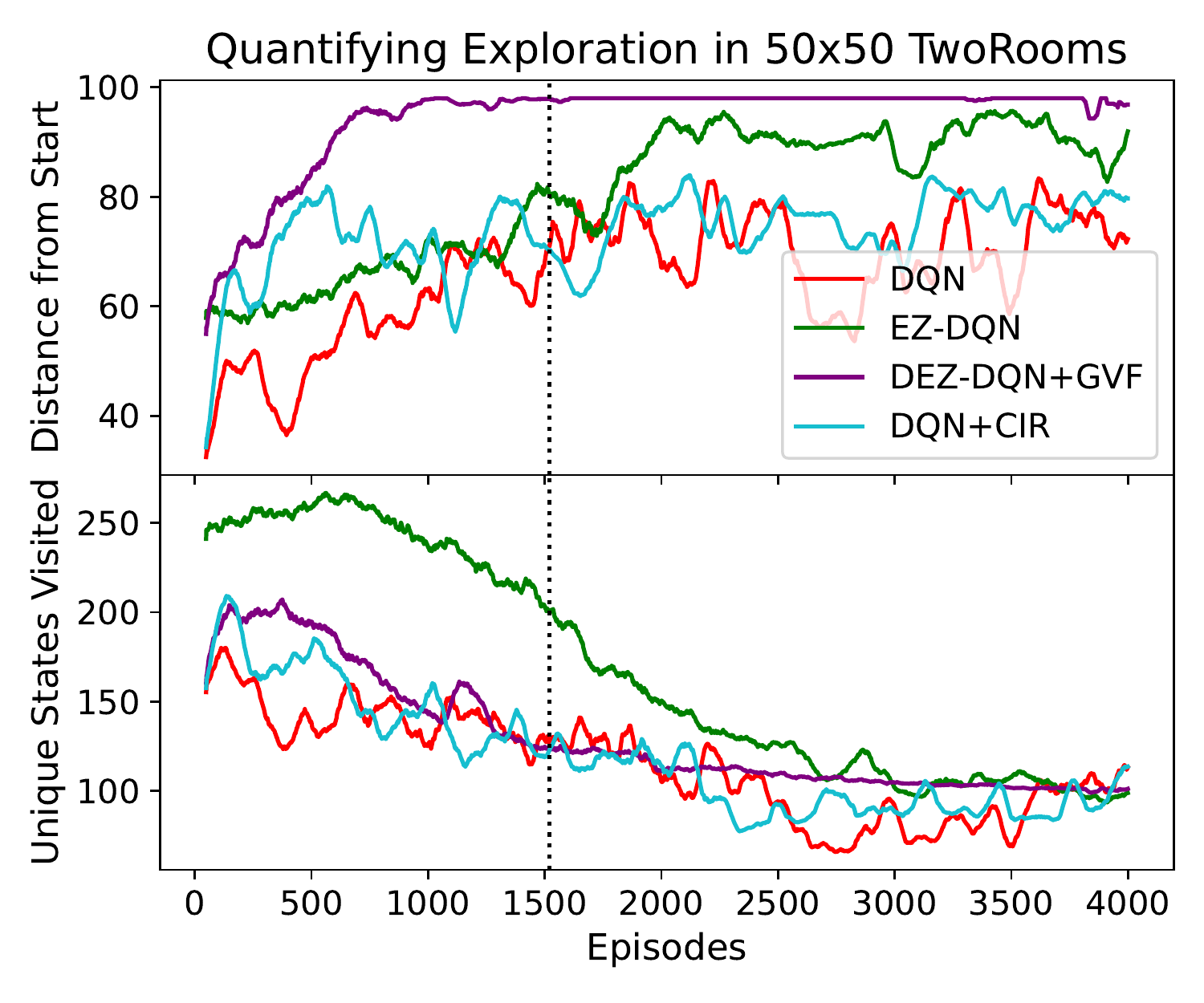}
	\caption{This plot tries to quantify exploration by comparing distance and unique states visited by normal epsilon greedy and directed epsilon greedy method. The graph is plotted with aggregating values for 100 episodes. }
	\label{fig:exploration1}
\end{figure*}

The Figure \ref{fig:exploration1} shows comparison between different exploration strategies in terms of the number of unique states visited and the farthest distance from the initial agent position for $8 \times 8$ DoorKey environments and $50 \times 50$ two rooms environments. Vertical line indicates the episode when DEZ-DQN+GVF reached $90\%$ of its convergent value.

\section{}
All the necessary hyper-parameters required for the necessary reproduction of the experiments are mentioned in the following table.
\paragraph{Note on DQN variants:} For the Two Rooms and SubGoal Two Rooms environment, we used a vanilla DQN with target networks and experience replay. The DQN variants used in DoorKey have an additional soft updating parameter, $\tau$ which basically keeps the target network from changing too much. At every target network update, we perform a weighted update of the target network parameters, with the weighting on the current train network weights being $\tau$. 

Also, for the MiniGrid environments, the weights were not updated at every time step, but after every 10 time-steps, which explains the high batch size used in these domains. These all modifications were used to get DQN to work properly for these environments, in an acceptable number of time-steps, making the use of as much parallel compute as possible.
\begin{figure*}[ht]
	\begin{center}
		\captionof{table}{Hyper-Parameters} 
		\label{table:parameters}
		\begin{tabular}{ | m{2cm} | m{7em}| m{4.5cm} | m{6cm} |}
			\hline
			Environment & Algorithm & Learning Parameters &  Model Parameters\\  
			\midrule
			TwoRooms 50x50 & 
			\pbox{20 cm}{DQN \\ DQN+GVF \\ EZDQN \\ EZDQN+GVF \\ DEZDQN+GVF \\ DQN+RS \\DQN+CIR} &
			\pbox{20cm}
			{
				runs =  5 \\
				episodes = 4000 \\
				batch = 4096 \\
				$\gamma$ = 0.99 \\
				$\alpha$ = 0.001 \\
				$\epsilon_{start}$ = 1.0 \\
				$\epsilon_{stop}$ =  0.01 \\ 
				$\epsilon_{decay}$ = $(1e-2)^{\frac{1}{episodes}}$ \\ 
				Persistence, $Z_{max}$ = 10 \\
				GVFs = 1\\
				$\eta$ = 0.01}
			&
			\pbox{20cm}{
				Architecture = [32 units, 32 units] \\
				Replay Buffer Size = 50000 \\
				Target Network Update =  100 steps \\
				Policy Network Update =  10 steps \\
				seeds = [54334, 82654, 21198,\\83554, 29948]
			}
			\\ 
			\midrule
			SubGoal TwoRooms 50x50 & \pbox{20 cm}{DQN \\ DQN+GVF \\ EZDQN \\ EZDQN+GVF \\ DEZDQN+GVF \\ DQN+RS \\DQN+CIR} &
			\pbox{20cm}
			{
				runs = 5 \\
				episodes = 8000 \\
				batch = 4096 \\
				$\gamma$ = 0.99 \\
				$\alpha$ = 0.001 \\
				$\epsilon_{start}$ = 1.0 \\
				$\epsilon_{stop}$ =  0.01 \\ 
				$\epsilon_{decay}$ = $(1e-3)^{\frac{1}{episodes}}$ \\ 
				Persistence, $Z_{max}$ = 30 \\
				GVFs = 2\\
				$\eta$ = 0.01}
			&
			\pbox{20cm}{
				Architecture = [32 units, 32 units] \\
				Replay Buffer Size = 60000 \\
				Target Network Update =  100 steps \\
				Policy Network Update =  10 steps \\
				seeds = [54334, 82654, 21198,\\83554, 29948]
			} \\
			\midrule
			6x6 DoorKey MiniGrid & \pbox{20 cm}{DQN \\ DQN+GVF \\ EZDQN \\ EZDQN+GVF \\ DEZDQN+GVF \\ DQN+RS \\DQN+CIR}   &
			\pbox{20cm}
			{
				runs = 5 \\
				episodes = 10000 \\
				batch = 2048 \\
				$\gamma$ = 0.95 \\
				$\alpha$ = 0.0001 \\
				$\tau$ = 0.05 \\
				$\epsilon_{start}$ = 1.0 \\
				$\epsilon_{stop}$ =  0.01 \\ 
				$\epsilon_{decay}$ = $(1e-2)^{\frac{1}{episodes \times \beta}}$ \\ 
				$\beta=0.5$\\
				Persistence, $Z_{max}$ = 3 \\
				GVFs = 2\\
				$\eta$ = 0.5}
			&
			\pbox{20cm}{
				Architecture: \\
				Conv1(C=16,F=2); \\
				MaxPool1(F=2,S=2); \\
				3 FC layer [60,60,6] \\
				Replay Buffer Size = 50000 \\
				Target Network Update =  1000 steps \\
				Policy Network Update =  10 steps \\
				seeds = [54334, 82654, 21198,\\83554, 29948]
			} \\
			\midrule
			8x8 DoorKey MiniGrid & \pbox{20 cm}{DQN \\ DQN+GVF \\ EZDQN \\ EZDQN+GVF \\ DEZDQN+GVF \\DQN+RS \\DQN+CIR } &
			\pbox{20cm}
			{
				runs = 5 \\
				episodes = 30000 \\
				batch = 4096 \\
				$\gamma$ = 0.95 \\
				$\alpha$ = 0.0001 \\
				$\tau$ = 0.05 \\
				$\epsilon_{start}$ = 1.0 \\
				$\epsilon_{end}$ =  0.01 \\ 
				$\epsilon_{decay}$ = $(1e-2)^{\frac{1}{episodes \times \beta}}$ \\ 
				$\beta=0.5$\\
				Persistence, $Z_{max}$ = 3 \\
				GVFs = 2\\
				$\eta$ = 0.5}
			&
			\pbox{20cm}{
				Architecture: \\
				Conv1(C=16,F=2); \\
				MaxPool1(F=2,S=1); \\ Conv2(C=16,F=2); \\
				MaxPool2(F=2,S=1); \\ 4 FC layer [120,60,10,6] \\
				Replay Buffer Size = 60000 \\
				Target Network Update =  1000 steps\\
				Policy Network Update =  10 steps \\
				seeds = [54334, 82654, 21198, \\83554, 29948] 
			} \\
			\midrule
		\end{tabular}
		
	\end{center}
\end{figure*}

\end{document}